\documentclass{article}

\usepackage{microtype}
\usepackage{graphicx}
\usepackage{subcaption}
\usepackage{booktabs} 

\usepackage{hyperref}

\usepackage[accepted]{icml2024}

\usepackage{amsmath}
\usepackage{amssymb}
\usepackage{mathtools}
\usepackage{amsthm}

\usepackage[capitalize,noabbrev]{cleveref}

\theoremstyle{plain}

\theoremstyle{definition}

\theoremstyle{remark}

\usepackage[textsize=tiny]{todonotes}

\newcommand*\modelname{Auto-regressive Diffusion}
\newcommand*\dataname{\textsc{MIS}}
\newcommand*\longmodelname{Many-to-many Diffusion}
\newcommand*\shortmodelname{M2M}

\newcommand*\sdmodel{M2M with Self-encoder}
\newcommand*\shortsdmodel{M2M-Self}

\newcommand*\dinomodel{M2M with DINO encoder}
\newcommand*\shortdinomodel{M2M-DINO}
\newcommand*\multiimageattn{Image-Set Attention}

\NewDocumentCommand{\ying}{ mO{} }{\textcolor{teal}{\textsuperscript{\textit{Ying}}\textsf{\textbf{\small[#1]}}}}

\icmltitlerunning{Many-to-many Image Generation with Auto-regressive Diffusion Models}

\begin{document}

\twocolumn[
\icmltitle{Many-to-many Image Generation with Auto-regressive Diffusion Models}


\icmlsetsymbol{equal}{*}

\begin{icmlauthorlist}
\icmlauthor{Ying Shen}{vt}
\icmlauthor{Yizhe Zhang}{apple}
\icmlauthor{Shuangfei Zhai}{apple}
\icmlauthor{Lifu Huang}{vt}
\icmlauthor{Joshua Susskind}{apple}
\icmlauthor{Jiatao Gu}{apple}
\end{icmlauthorlist}

\icmlaffiliation{apple}{Apple, USA}
\icmlaffiliation{vt}{Computer Science Department, Virginia Tech, Blacksburg, USA}

\icmlcorrespondingauthor{Ying Shen}{yings@vt.edu}
\icmlcorrespondingauthor{Jiatao Gu}{jiatao@apple.com}

\icmlkeywords{Machine Learning, ICML}

\vskip 0.3in
]



\printAffiliationsAndNotice{}  

\begin{abstract}


Recent advancements in image generation have made significant progress, yet existing models present limitations in perceiving and generating an arbitrary number of interrelated images within a broad context. 
This limitation becomes increasingly critical as the demand for multi-image scenarios, such as multi-view images and visual narratives, grows with the expansion of multimedia platforms.
This paper introduces a domain-general framework for many-to-many image generation, capable of producing interrelated image series from a given set of images, offering a scalable solution that obviates the need for task-specific solutions across different multi-image scenarios.
To facilitate this, we present \dataname{}, a novel large-scale multi-image dataset, containing 12M synthetic multi-image samples, each with 25 interconnected images. Utilizing Stable Diffusion with varied latent noises, our method produces a set of interconnected images from a single caption.
Leveraging \dataname{}, we learn \shortmodelname{}, an autoregressive model for many-to-many generation, where each image is modeled within a diffusion framework. 
 Throughout training on the synthetic \dataname{}, the model excels in capturing style and content from preceding images — synthetic or real — and generates novel images following the captured patterns.
Furthermore, through task-specific fine-tuning, our model demonstrates its adaptability to various multi-image generation tasks, including Novel View Synthesis and Visual Procedure Generation.

\end{abstract}    
\section{Introduction}
\label{sec:intro}

\begin{figure}[!t]
    \centering
\includegraphics[width=\linewidth]{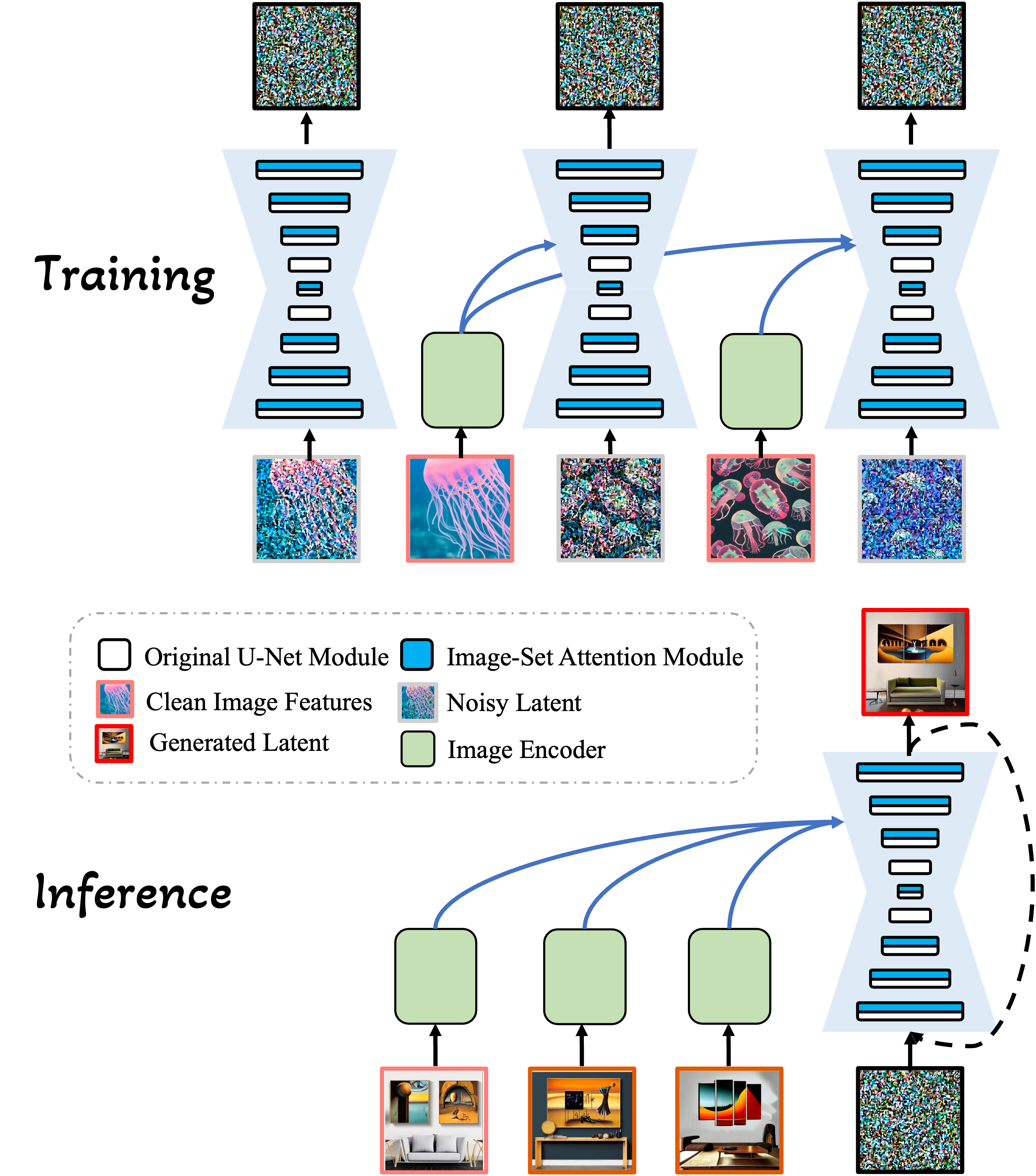}
\vspace{-5pt}
    \caption{\textbf{Overview of \modelname{} Pipeline.} During training, \modelname{} takes sets of noised latent images and their corresponding clean image features as inputs and then predicts the noise added to each noised latent image conditioned on the previous clean image features. At inference, our model can generate an arbitrary number of images in an auto-regressive manner by iteratively incorporating generated images back into the input. 
}
    \label{fig:overview}
    \vspace{-5mm}
\end{figure}

Over the past few years, the realm of image generation has seen remarkable progress across numerous tasks, including super-resolution~\cite{ho2022cascaded,saharia2022image}, image manipulation~\cite{meng2021sdedit,nichol2022glide,kawar2023imagic,brooks2023instructpix2pix},  text-to-image generation~\cite{ramesh2021zero,rombach2022high,ramesh2022hierarchical,saharia2022photorealistic,yu2022scaling}, and more. 
Despite this progress, the primary focus has predominantly been on the generation or processing of individual images.
Some studies \cite{li2019storygan,maharana2022storydall,pan2022synthesizing,liu2023intelligent} explore beyond single image generation but usually with a specific focus on specific tasks like story synthesis~\cite{li2019storygan}, visual in-context learning~\cite{bar2022visual}, novel-view synthesis~\cite{liu2023zero}, and so on.
These examples illustrate the versatility of the multi-image generation paradigm, which can also encompass video generation, essentially a sequence of image frames.
The proliferation of multimedia platforms has increased the need for accepting multiple images and generating multiple images that possess different types of interconnection, catering to applications like producing thematically and stylistically consistent image sets for advertising or displaying objects from various perspectives. 
Recognizing the merging demand, this paper thus underscores the need for a holistic exploration into the general-domain multi-image to multi-image generation paradigm, where models are designed to perceive and generate an arbitrary number of interrelated images within a broader context.

In this work, we present a domain-general framework for multi-image to multi-image generation that can perceive and generate a flexible number of interrelated images auto-regressively, thus offering the flexibility and adaptability needed to meet a broad range of multi-image generation tasks.
A cornerstone of this endeavor is the exposure of our framework to a diverse collection of multi-image examples that inherently maintain meaningful interrelations amongst the images in each set. 
To facilitate this, we introduce \dataname{}, the first large-scale multi-image dataset comprising sets of images interconnected by general semantic relationships. 
Unlike previous multi-image datasets specialized towards specific scenarios, such as sequential frames or images from multiple viewpoints, \dataname{} encapsulates more general semantic interconnections among images. 
\dataname{} consists of a total of 12M synthetic multi-image set samples, each containing 25 interconnected images. 
Motivated by the success of diffusion models in text-to-image generation~\cite{rombach2022high,ramesh2022hierarchical,saharia2022photorealistic,nichol2022glide,yu2022scaling}, we propose to utilize Stable Diffusion to produce interconnected image sets from identical caption but varied latent noise, ensuring coherence and uniqueness within each set.

Leveraging our \dataname{} dataset, we propose \longmodelname{} (\shortmodelname{}), a conditional diffusion model that can perceive and generate an arbitrary number of interrelated images in an auto-regressive manner. 
\shortmodelname{} excels in generating sequences of interconnected images from purely visual inputs, without reliance on textual descriptions.
\shortmodelname{} is built on top of latent diffusion models and we extend it into a multi-image to multi-image generator by introducing an Image-Set Attention module that learns to capture the intricate interconnections within a set of images, thereby facilitating more contextually coherent multi-image generation.
Our study further explores various architectural designs for multi-image to multi-image generation tasks, focusing on diverse strategies for handling the preceding images. As part of our contributions, we introduce two novel model variants: \sdmodel{} (\shortsdmodel{}) and \dinomodel{} (\shortdinomodel{}). \shortsdmodel{} utilizes the same U-Net-based denoising model to simultaneously process preceding latent images along with noisy latent images, enabling more refined cross-attention.
Meanwhile, \shortdinomodel{} employs external vision models to encode preceding images, leveraging the power of more discriminative visual features.
Experimental results demonstrated that our proposed method learns to capture style and content from preceding images and generate novel images in alignment with the observed patterns. Impressively, despite being trained solely on synthetic data, our model exhibits zero-shot generalization to \textit{real} images. 
Furthermore, through task-specific fine-tuning, our model demonstrates its adaptability to various multi-image generation tasks, including Novel View Synthesis and Visual Procedure Generation, suggesting its potential to handle complex multi-image generation challenges.

Our paper makes the following contributions: 

(1) We introduce an innovative strategy for constructing \dataname{}, the first large-scale multi-image dataset containing 12M synthetic multi-image set samples, each with 25 images interconnected by general semantic relationships.

(2) We propose a domain-general \longmodelname{} (\shortmodelname{}) model that can perceive and generate an arbitrary number of interrelated images in an auto-regressive manner.

(3) We demonstrate that \shortmodelname{} learns to capture style and content from preceding images and generate novel images following the captured patterns. 
It exhibits great zero-shot generalization to real images and offers notable potential for customization to specific multi-image generation tasks.
\section{Related Work}
\label{sec:related}

\subsection{Image Generation}

Image generation has always been a heated topic in the field of computer vision. demand .. in many different generation tasks, such as super-resolution~\cite{ho2022cascaded,saharia2022image}, image manipulation~\cite{meng2021sdedit,nichol2022glide,kawar2023imagic,brooks2023instructpix2pix},  text-to-image generation~\cite{ramesh2021zero,rombach2022high,ramesh2022hierarchical,saharia2022photorealistic,yu2022scaling}, and so on. 
Beyond the realm of single-image generation, there has been an exploration into multi-image generation~\cite{li2019storygan,bar2022visual,liu2023zero} but usually with a specific focus on specific tasks.
For instance, story synthesis~\cite{li2019storygan} aims at generating a series of images that narrate a coherent narrative. Other tasks, such as visual in-context learning~\cite{bar2022visual} focus on generating one target image using a query image accompanied by example image pairs. Additionally, novel-view synthesis~\cite{liu2023zero} aims to generate images from new viewpoints based on a set of posed images of a particular scene or object. 
Our work distinctively expands on the existing literature by proposing a more general multi-image to multi-image generation framework, that can be adapted to a myriad of scenarios in multi-image generation.

\subsection{Diffusion-based Generative Models}

Diffusion-based generative models\cite{ho2020denoising,song2019generative} have marked notable progress in the field of generative AI. These models operate by incrementally introducing noise to perturb the data and learning to generate data samples from Gaussian noise through a series of de-noising processes.
Recently, diffusion models have become prominent for generating images~\cite{rombach2022high,ramesh2022hierarchical,saharia2022photorealistic,brooks2023instructpix2pix}, as well as in other modalities such as video~\cite{ho2022imagen,singer2022make,blattmann2023align}, audio~\cite{pmlr-v202-liu23f,pmlr-v202-huang23i}, text~\cite{li2022diffusion,gong2022diffuseq,zhang2023planner}, 3D~\cite{poole2022dreamfusion,gu2023nerfdiff,liu2023zero} and more. Given their superior performance in image generation, we propose to leverage the diffusion-based generative models for two key purposes: constructing the synthetic multi-image set dataset, \dataname{}, and serving as the backbone architecture for our multi-image generation framework. 

\section{Background}
\label{sec:background}
\begin{figure}[!t]
    \centering
\includegraphics[width=\columnwidth]{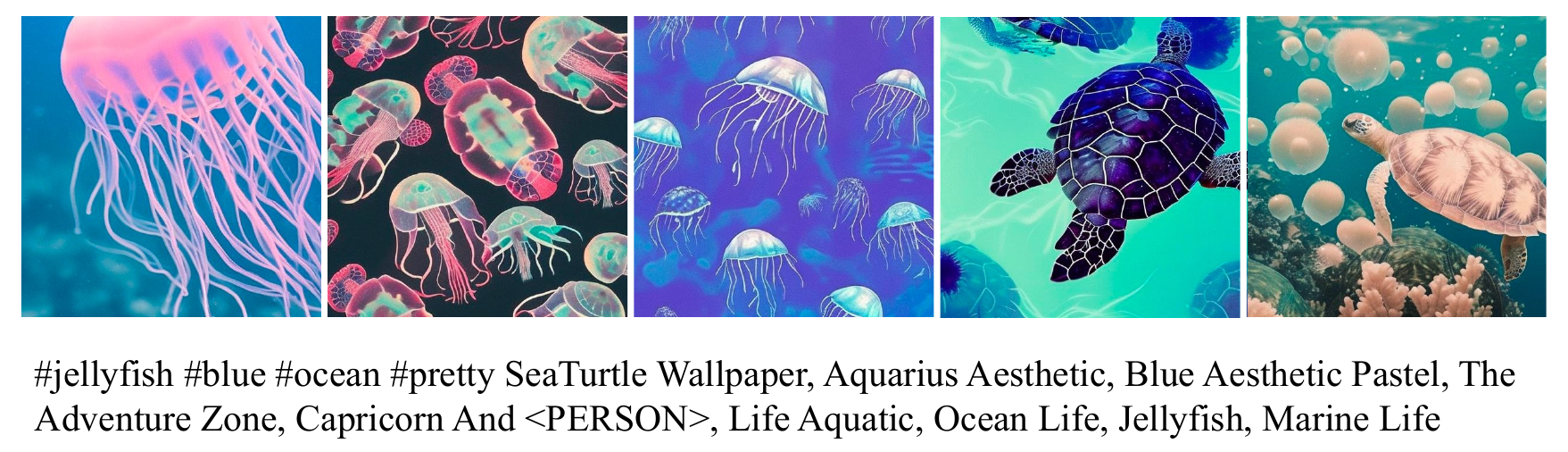}
    \caption{\textbf{A sample image set of five distinctive images generated using a caption from Conceptual 12 M.}}
    \label{fig:multi_image_sample}
    \vspace{-15pt}
\end{figure}
The Diffusion Probabilistic Models \cite{ho2020denoising,song2019generative} learns the data distribution $p(x)$ by gradually denoising a normally distributed variable throughout a Markov chain with length $T$. Specifically, diffusion models define a forward diffusion process in a Markov chain, incrementally adding Gaussian noise samples to the initial data point $x$ into the Gaussian noise $x_T$ over $T$ steps, and a learnable reverse process that denoises $x_T$ back to the clean input $x$ iteratively via a sequence of time-conditioned denoising autoencoders $\epsilon_{\theta}(x_t, t)$. Typically, the denoising model $\epsilon_{\theta}$ is implemented via time-conditioned U-Net \cite{ronneberger2015u}.
The diffusion model is commonly trained with a simplified L2 denoising loss \cite{ho2020denoising}:
\begin{align}
    \mathcal{L}_{DM} = \mathbb{E}_{x, \epsilon, t} \big{[} \| \epsilon - \epsilon_{\theta}(x_t, t) \|^2_2 \big{]}
\end{align}
where $\epsilon \sim \mathcal{N}(0, 1)$, $t \sim \mathcal{U}(0, T)$. 

\paragraph{Latent Diffusion Models} Latent Diffusion Models \cite{rombach2022high} improves the efficiency of diffusion models by operating in the latent representation space of a pretrained variational autoencoder \cite{kingma2013auto} with encoder $\mathcal{E}$ and decoder $\mathcal{D}$, such that $\mathcal{D}(\mathcal{E}(x)) \approx x$. 

Diffusion models can be conditioned on various signals such as class labels or texts. The conditional latent diffusion models learn a denoising model $\epsilon_{\theta}$ that predicts the noise added to the noisy latent $z_t$ given conditioning $c$ via the following objective:
\begin{align}
    \mathcal{L}_{LDM} = \mathbb{E}_{\mathcal{E}(x), c, \epsilon, t} \big{[} \| \epsilon - \epsilon_{\theta}(z_t, t, c) \|^2_2 \big{]}
\end{align}

During inference, classifier-free guidance \cite{ho2021classifier} is employed to improve sample quality:
\begin{align}
    \tilde{\epsilon}_{\theta}(z_t, c) = \epsilon_{\theta}(z_t, \varnothing) + s \cdot (\epsilon_{\theta}(z_t, c) - \epsilon_{\theta}(z_t, \varnothing))
\label{eq:sample}
\end{align}

where $\epsilon_{\theta}(z_t, c)$ and $\epsilon_{\theta}(z_t, \varnothing)$ refer to the condition and unconditional $\epsilon$-predictions, and $s$ represents the guidance scale. Setting $s = 1$ disables the classifier-free guidance while increasing $s > 1$ strengthens the effect of guidance. 

\section{Multi-Image Set Construction}
\label{sec:dataset}

Learning to process and generate multiple images requires a diverse collection of multi-image examples for training.
As manually collecting a multi-image dataset comprising sets of interconnected images would be resource-intensive, we propose to leverage the capabilities of text-to-image models (Latent Diffusion Model \cite{rombach2022high}) for generating a multi-image dataset, which we refer to as \dataname{}. This dataset comprises sets of interconnected images. Specifically, we leverage the power of the Latent Diffusion Model and its capacity to generate a diverse set of images from the same caption by employing different latent noises. This approach allows us to construct a set of interconnected images for each caption, recognizing that these images within the set exhibit internal connections owing to the common caption they were generated with. Notably, while all images generated from the same caption are drawn from the same conditional probability distribution, underscoring their meaningful internal connections, the introduction of varied latent noise ensures the distinctiveness of each image within the set.

To gather image captions, we employ Conceptual 12M \cite{changpinyo2021conceptual}, a large image-text pair dataset that contains approximately 12 million web images, each accompanied by corresponding a descriptive alt-text. For \dataname{} generation, we exclusively utilize these alt-texts as captions for generating interconnected images. More precisely, with each caption, we employ the Stable Diffusion model to generate a set of distinct images by using different latent noise.
We utilize Stable Diffusion v2-1-base \footnote{https://huggingface.co/stabilityai/stable-diffusion-2-1-base} along with the Euler Discrete Scheduler \cite{karras2022elucidating} and a default guidance scale of 7.5 for image generation. Stable Diffusion generates images conditioned on CLIP \cite{radford2021learning} text embeddings. To ensure compatibility with the maximum token length allowed for the CLIP Text encoder, we filter out captions exceeding the maximum token length of 77 tokens. This results in a curated collection of 12,237,187 captions for our multi-image generation. For each individual caption, we employ different latent noises to generate 25 distinct images. As a result, our final \dataname{} consists of 12M multi-image set samples, with each image set containing 25 interconnected images. Figure~\ref{fig:multi_image_sample} shows an example of an image set generated using a single caption from Conceptual 12M.

\section{\longmodelname{} (\shortmodelname{})}
\label{sec:method}

We introduce the \longmodelname{} (\shortmodelname{}) framework, designed to perceive and generate an arbitrary number of interrelated images auto-regressively, as illustrated in Figure \ref{fig:overview}. 
Our framework extends the pre-trained Stable Diffusion, the large-scale text-to-image latent diffusion model. Central to Stable Diffusion is the denoising model $\epsilon_{\theta}(\cdot)$, which is built upon the U-Net, but further enriched with a text-to-image cross-attention layer. 
We modify the architecture by supplanting the text-to-image cross-attention module with our \multiimageattn{} module, which allows the model to learn and understand the intricate interconnections within a set of images, thereby facilitating more contextually coherent multi-image generation.

\shortmodelname{} explores various architectural approaches for multi-image generation, with a focus on how preceding images are encoded. We discuss two main model variants: the \textbf{\sdmodel{} (\shortsdmodel{})} in Section \ref{sec:sdmodel} and the \textbf{\dinomodel{} (\shortdinomodel{})} in Section \ref{sec:dinomodel}.
\shortsdmodel{} leverages the U-Net-based denoising model for simultaneously processing the preceding and the noisy latent images, enabling the cross-attention mechanisms over various spatial dimensions of the preceding images. 
Meanwhile, \shortdinomodel{} explores integrating external vision models for encoding preceding images, aiming to complement the U-Net's inherent capabilities for encoding preceding images.


\subsection{\sdmodel{} (\shortsdmodel{})}
\label{sec:sdmodel}

During the training phase, given an image set $\{I\}_{i=1}^N$, each image $I_i$ is first encoded individually into the latent code $z_0^i$ using a pre-trained autoencoder $z_0^i = \mathcal{E}(I_i)$. These image latents are then stacked to construct $z_0^{1:N} \in \mathbb{R}^{N \times C \times H \times W}$, where $N$ represents the number of images within each set, $C$ is the number of latent channels, and $H$ and $W$ are the spatial dimensions of the latent space.
The clean latent $z_0^{1:N}$ is subsequently noised according to the pre-defined forward diffusion schedule to produce the noisy latent $z_t^{1:N}$, where the noise level increases over diffusion timesteps $t$. 

To prepare the input for \shortsdmodel{}, we concatenate the clean and noisy latent codes, resulting in $z^{1:N} = [z_0^{1:N}; z_t^{1:N}] \in \mathbb{R}^{2 \times N \times C \times H \times W}$. Here, the symbol $;$ denotes the concatenation operation and the factor $2$ indicates the inclusion of both clean and noised latent forms.
\shortsdmodel{} thus takes the concatenated tensor $z^{1:N}$ as inputs and predicts the noise added to each noised latent image in an auto-regressive manner.

\begin{figure}[!t]
    \centering
\includegraphics[width=0.8\columnwidth]{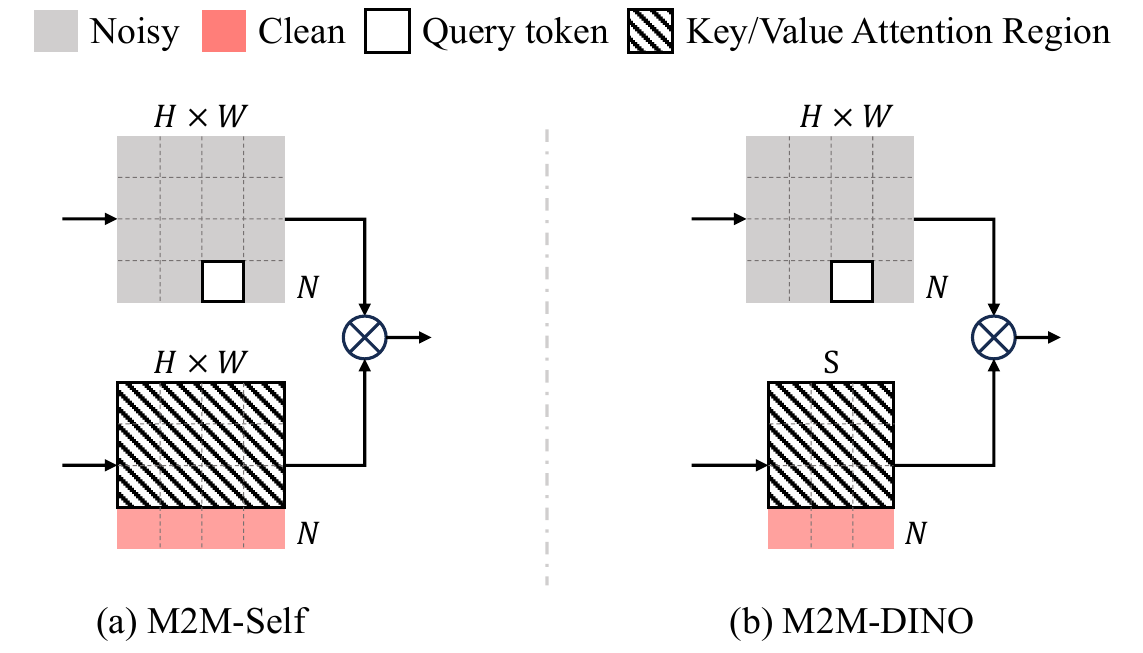}
    \caption{\textbf{Illustration of \multiimageattn{} Module.} The query token is denoted in a white square and its corresponding key/value attention region is marked by a diagonal striped pattern.}
    \label{fig:global}
    \vspace{-5mm}
\end{figure}

\paragraph{\multiimageattn{}}

The core component in \shortsdmodel{} is the \multiimageattn{} module, designed to facilitate effective cross-attention from noisy latent images to their preceding clean latent images.
The \multiimageattn{} module accepts an input tensor $z \in \mathbb{R}^{BZ \times 2 \times N \times H \times W \times C}$, where $BZ$ stands for batch size. For readability, the superscript $1:N$ has been omitted. 
The input $z$ is first partitioned and reshaped into the following two components: the noisy latent $z_n \in \mathbb{R}^{BZ \times (N \times H \times W) \times C}$ and its corresponding clean latent $z_c \in\mathbb{R}^{BZ \times (N \times H \times W) \times C}$.  This transformation enables a comprehensive cross-attention across the resultant length of $N \times H \times W$. 
These two latents are then projected and processed through the scaled dot-product cross-attention layer, as introduced by \cite{vaswani2017attention}:
\begin{align}
    z'_n = \text{Attention}(Q, K, V) = \text{Softmax}(\frac{QK^T}{\sqrt{d}}) \cdot V,
\label{eq:attn}
\end{align}
where $Q = W^Q z^Q$, $K = W^K z^K$, and $V = W^V z^V$ denotes the projections of the reshaped latents, with $d$ indicating latent feature dimension.
The cross-attention operates with the noisy latents as queries $z^Q = z_n$ and the clean latents as keys and values $z^K = z^V = z_c$.
We also make the cross-attention multi-head, allowing the model to jointly attend to information from different representation subspace~\cite{vaswani2017attention}.
To maintain the flow of information from previous clean images only, a causal image-set attention mask is introduced, restricting each noisy latent $z^i_{n}$ to exclusively attend to patches within its preceding clean latents $z^{<i}_c$. 
Figure \ref{fig:global}(a) provides a visualization of this process, depicting an example of a query token and the key/value region it allows to attend.
The output latent $z'$ is formed by concatenating the original clean latent and the updated noisy latent, resulting in $z' = [z_c; z'_n]$.

\subsection{\dinomodel{} (\shortdinomodel{})}
\label{sec:dinomodel}

In addition to the \shortsdmodel{}, which primarily relies on the same U-Net-based denoising model for processing the preceding images as internal features, we explore the potential advantage of integrating external vision models to enhance the encoding of preceding images into more discriminative visual features. Specifically, we leverage DINOv2~\cite{oquab2023dinov2}, a model renowned for its superior performance in understanding fine-grained vision information.

\paragraph{\multiimageattn{}}
In \shortdinomodel{}, each image $I_i$ from a set $\{I\}_{i=1}^N$ is encoded into two distinct formats: the latent code $z_0^i$ using a pre-trained autoencoder $z_0^i = \mathcal{E}(I_i)$ and DINO features $v^i = \mathcal{E}_v(I_i)$ with the DINO image encoder $\mathcal{E}_v(\cdot)$. 
These encoded forms are then stacked separately to construct two forms of features: $z_0^{1:N} \in \mathbb{R}^{N \times C \times H \times W}$ and $v^{1:N} \in \mathbb{R}^{N \times S \times D}$, where $N$ represents the number of images within each set, $S$ the length of DINO image tokens, and $D$ the dimensionality of the DINO features.
Subsequently, noisy latents $z_t^{1:N}$ are constructed from $z_0^{1:N}$ according to a predefined noise addition schedule. For clarity, the superscript $1:N$ will be omitted in this context.

The \multiimageattn{} module in \shortdinomodel{} takes in the noisy latent $z_t \in \mathbb{R}^{BZ \times N \times H \times W \times C}$ and the DINO features of clean images $v \in \mathbb{R}^{BZ \times N \times S \times W}$. It reshapes $z_t$ into $\mathbb{R}^{BZ \times (N \times H \times W) \times C}$ and $v$ into $\mathbb{R}^{N \times (N \times S) \times D}$ to facilitate the cross-attention.
The cross-attention mechanism, as detailed in Equation \ref{eq:attn}, employs DINO features $v$ act as keys and values $z^K = z^V = v$ and the noisy image latents $z_t$ as queries $z^Q$, facilitating interaction between noisy and clean features as depicted in Figure \ref{fig:global}(b).
Additionally, akin to \shortsdmodel{}, \shortdinomodel{} utilizes a causal image-set attention mask, permitting each noisy latent $z^i_{t}$ to exclusively attend to tokens within its preceding DINO features $v^{<i}$.

\section{Training and Inference Procedure}

\subsection{Initial Training and Inference}
\paragraph{Training Objective} The training process is similar to the Latent Diffusion Model \cite{rombach2022high}. During training, we initialize the model weights from the Stable Diffusion Model, but intentionally exclude the pre-trained text-to-image cross-attention layer, and initialize our \multiimageattn{} module from scratch. 
\modelname{} takes the visual features of the clean images, denoted as $z_0^{1:N}$, and the noisy latent images, $z_t^{1:N}$, as inputs. The model's objective is to predict the noise strength added to each noised latent image conditioned on the previous clean image features. We employ an auto-regressive training strategy, where the loss is accumulated based on the difference in the predicted noise of each image, as follows:
\begin{align}
    \mathcal{L}_{\theta} = \mathbb{E}_{\mathcal{E}(x_0^{1:N}), \epsilon, t} \big{[} \| \epsilon - \epsilon_{\theta}(z_t^{1:N}, z_0^{1:N}, t) \|^2_2 \big{]},
\end{align}
where $z_0^{1:N}$ corresponds to the clean image latents in \shortsdmodel{} and DINO features of the clean images in \shortdinomodel{}.
To facilitate classifier-free guidance~\cite{ho2021classifier} during inference, \modelname{} is jointly trained with conditional and unconditional objectives, and the conditional and unconditional score estimates are combined at inference time. Training for unconditional denoising is achieved by randomly zeroing out the clean image condition $z_0^{1:N}$ with 10\% probability.

\paragraph{Inference}

As shown in Figure \ref{fig:overview}, during inference, \modelname{} exhibits the capability to auto-regressively generate multiple images. This is achieved by iteratively incorporating the images generated in the previous iteration as new inputs for subsequent iterations. This process begins by generating an initial image, given a set of conditional input images and randomly sampled noise latent codes. Afterward, the model incorporates the previously generated image into the input context for the next generation cycle. In practice, the number of context images that can be input into the model is bounded by a predefined parameter, the context window $W$.
At inference time, with a guidance scale $s \geq 1$, the output of the model is extrapolated further in the direction of the conditional $\epsilon_{\theta}(z_t, z_0^{1:W}, t)$ and away from the unconditional $\epsilon_{\theta}(z_t, \mathbf{0}, t)$:
\begin{align}
    \tilde{\epsilon}_{\theta}(z_t, z_0^{1:W}, t) &= \epsilon_{\theta}(z_t, \mathbf{0}, t) \notag \\
    &\quad + s \cdot (\epsilon_{\theta}(z_t, z_0^{1:W}, t) - \epsilon_{\theta}(z_t, \mathbf{0}, t))
    \label{eq:sample_general}
\end{align}

Here, $z_0^{1:W}$ represents the sequence of context images serving as the conditional inputs for the generative process.
\begin{figure*}[!t]
    \centering
    \begin{subfigure}[b]{0.47\linewidth}  
        \centering
        \includegraphics[width=\linewidth]{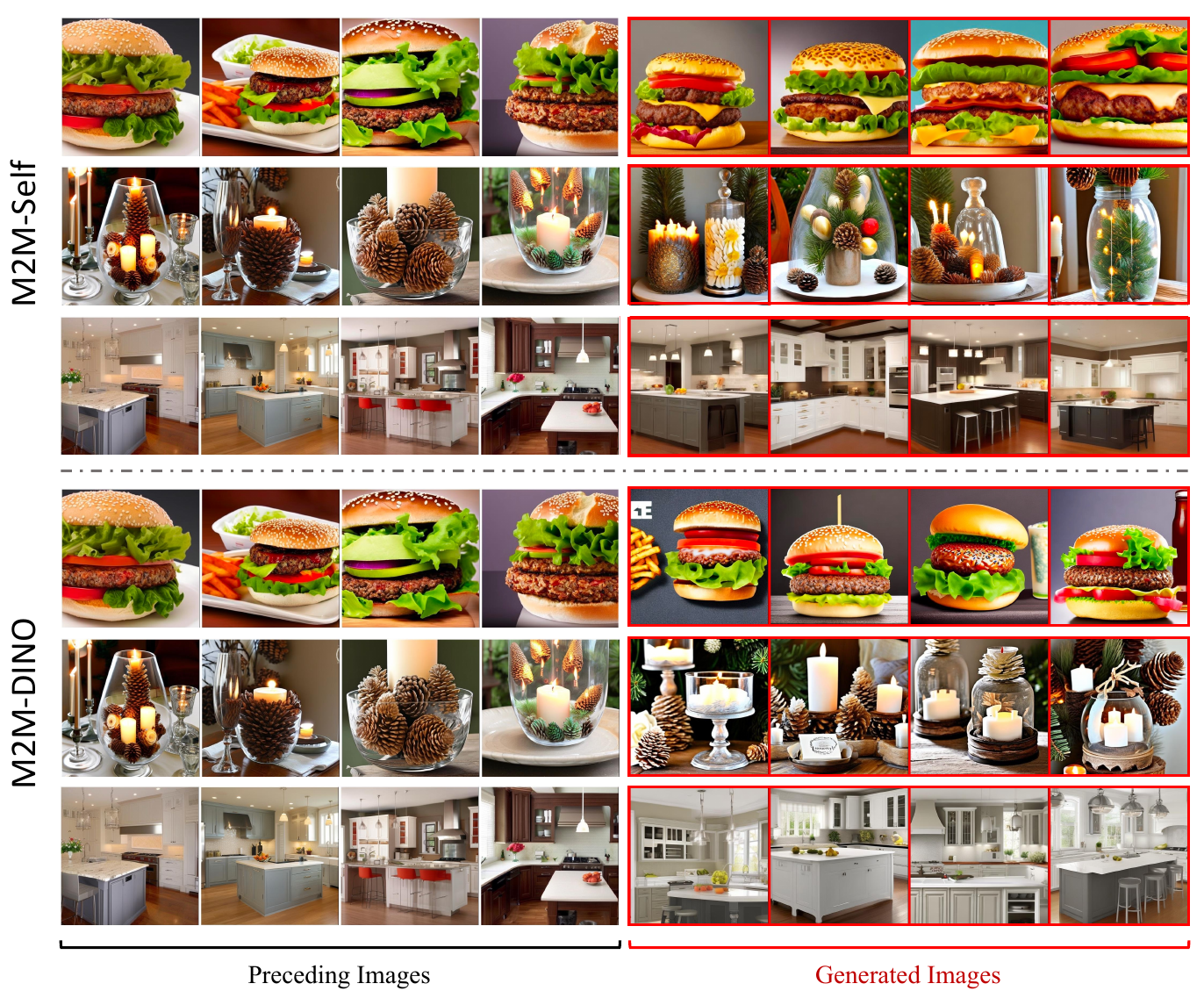}
        \caption{Content Consistency}
        \label{fig:content}
    \end{subfigure}
    \hspace{3mm}  
    \begin{subfigure}[b]{0.47\linewidth}  
        \centering
        \includegraphics[width=\linewidth]{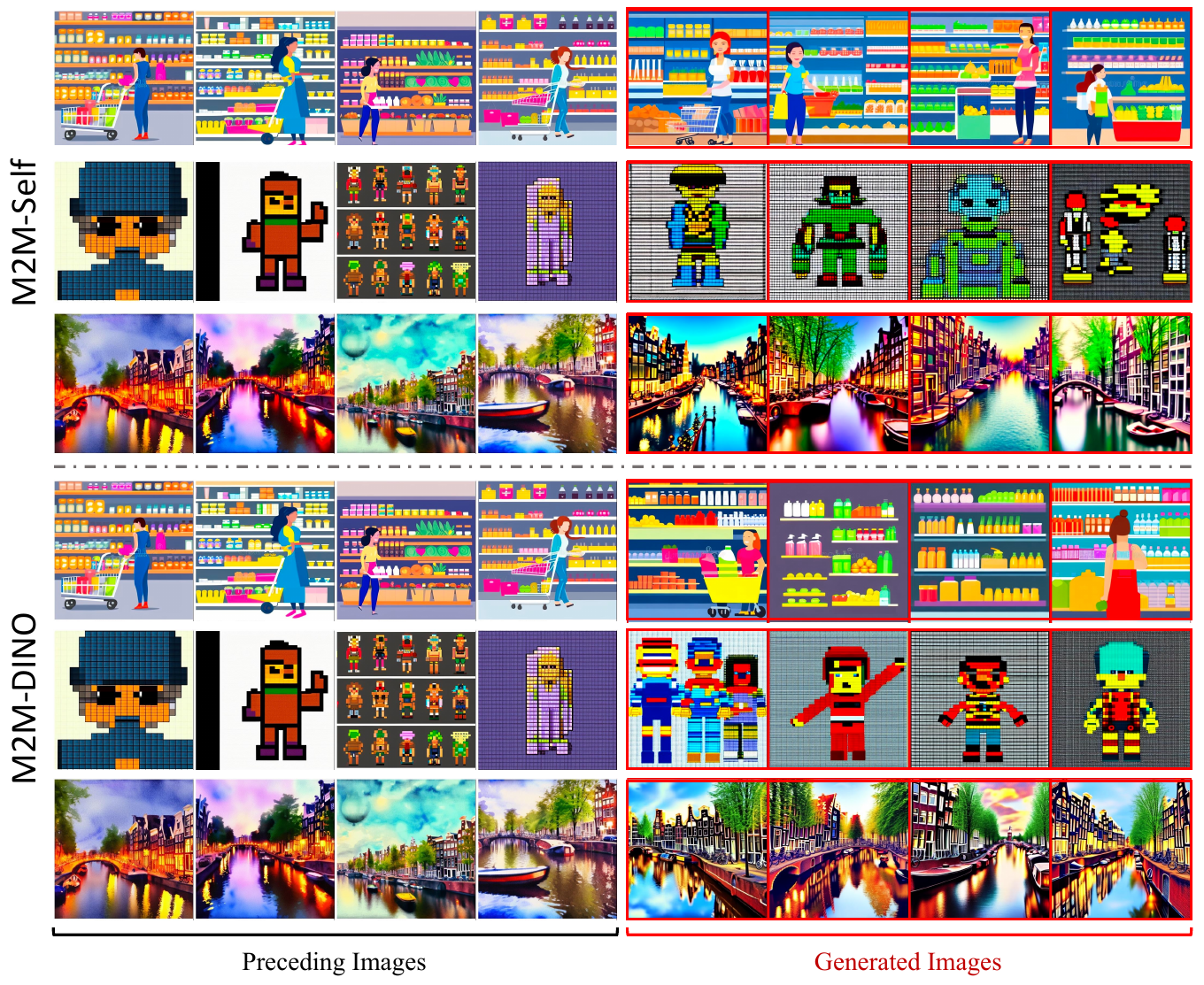}
        \caption{Style Consistency}
        \label{fig:style}
    \end{subfigure}
    \vspace{-1mm}
    \caption{\textbf{Consistency Evaluation in \shortsdmodel{} and \shortdinomodel{}}: 
    The figure showcases the ability of \shortsdmodel{} and \shortdinomodel{} to maintain content (a) and style (b) consistency. Content consistency refers to the model's capacity to generate images with the same type of subject as preceding ones, while style consistency pertains to maintaining aesthetic elements like color schemes, textures, and artistic techniques.
    Each subfigure contains two panels: the top panel for \shortsdmodel{} and the bottom for \shortdinomodel{}. Columns 1-4 showcase the preceding images for conditioning, and Columns 5-8 display images generated by the respective models.}
    \label{fig:consistency}
    \vspace{-5mm}
\end{figure*}

\subsection{Task-specific Fine-tuning and Inference}

\paragraph{Fine-tuning}

Building upon the initial training on \dataname{}, we extend \shortmodelname{}'s capabilities through task-specific fine-tuning for various multi-image generation tasks by introducing the task-specific conditions $c_C$ through additional embeddings.
Depending on the task, the conditions $c_C$ can vary, encompassing positional information, camera viewpoints, and others, tailored to capture the unique aspects of each specific task. These conditions are integrated into \shortsdmodel{} by adding them to the hidden states before applying \multiimageattn{}, formulated as $z^{1:N} = z^{1:N} + E(c_C)$, where $E(\cdot)$ represents a general embedding strategy that varies under different conditions $c_C$.
In the case of \dinomodel{}, where we utilize DINO features for encoding preceding images, a distinct conditional embedder, $E_c(\cdot)$, is applied, represented by $v^{1:N} = v^{1:N} + E_c(c_C)$.
This approach enables the model to incorporate additional contextual or spatial information relevant to the task, thereby enhancing its capability to generate images aligned with the task-specific requirements.
To facilitate classifier-free guidance with the task-specific conditions during sampling, we randomly zero out the task-specific conditions along with the clean image condition $z_0^{1:N}$ with 10\% probability. 


\paragraph{Inference}

During inference, \shortmodelname{} employs distinct guidance scales, $s_I$ for the clean image condition $z_0^{1:W}$ and $s_C$ for task-specific conditions $c_C$. This facilitates refined control over image generation, catering to both the image context and the specific requirements of the task at hand.
Consequently, Equation \ref{eq:sample_general} is adapted as follows:
\begin{align}
    \tilde{\epsilon}_{\theta}(&z_t, z_0^{1:W}, c_C) \nonumber\\
    &= \epsilon_{\theta}(z_t, \varnothing, \varnothing) \nonumber \\ 
    &+ s_I \cdot \big(\epsilon_{\theta}(z_t, z_0^{1:W}, \varnothing) - \epsilon_{\theta}(z_t, \varnothing, \varnothing)\big) \nonumber\\ 
    & + s_C \cdot \big(\epsilon_{\theta}(z_t z_0^{1:W}, c_C) - \epsilon_{\theta}(z_t, z_0^{1:W}, \varnothing)\big)
\end{align}

\section{Experimental Setup}
\label{sec:exp_setup}

\subsection{Datasets} 

\subsubsection{Pre-training Datasets}
For pre-training, we leverage our introduced \dataname{}, as detailed in Section~\ref{sec:dataset}. We adopt two distinct subsets of this dataset for training two model variants: \sdmodel{} (\shortsdmodel{}) and \dinomodel{} (\shortdinomodel{}).
Specifically, the \shortsdmodel{} model is trained on a subset of 9M multi-image examples, each containing a set of $N=5$ images. Meanwhile, \shortdinomodel{} is trained on a subset consisting of 6M multi-image examples.


\subsubsection{Task-Specific Fine-tuning Datasets}

\paragraph{Objaverse} Objaverse~\cite{deitke2023objaverse}, a large-scale dataset containing 800K+ 3D objects. Each object in the dataset contains 12 images of the object from different camera viewpoints along with the associated 12 camera poses. We utilize Objaverse to evaluate the model's performance in adapting to the novel view synthesis task. Given several posed images of a specific object, this task aims to generate images of the same object from novel viewpoints.

\paragraph{Visual Goal-Step Inference} Visual Goal-Step Inference (VGSI)~\cite{yang2021visual} comprises approximately 53K wikiHow articles across various categories of everyday tasks. Each wikiHow article contains one or more different methods to achieve it, with each method including a series of specific steps accompanied by corresponding images. We employ this dataset to construct the visual procedure generation task where the model learns to generate images depicting future steps given images from preceding steps.




\subsection{Evaluation Metrics}

To assess the performance of our models, we employ Fréchet Inception Distance (FID) \cite{heusel2017gans} and inception score (IS)~\cite{salimans2016improved} to measure the quality of generated images. Additionally, we utilize the CLIP score \cite{radford2021learning,hessel2021clipscore} to measure the alignment of the generated images with their preceding images. Our evaluation is conducted on a random selection of 10K samples from the test split of \dataname{}. Furthermore, we present both qualitative evaluations from this test split and real-world images.

\section{Results and Discussion}
\label{sec:results_diss}
\begin{figure}[!t]
    \centering
    \includegraphics[width=0.9\linewidth]{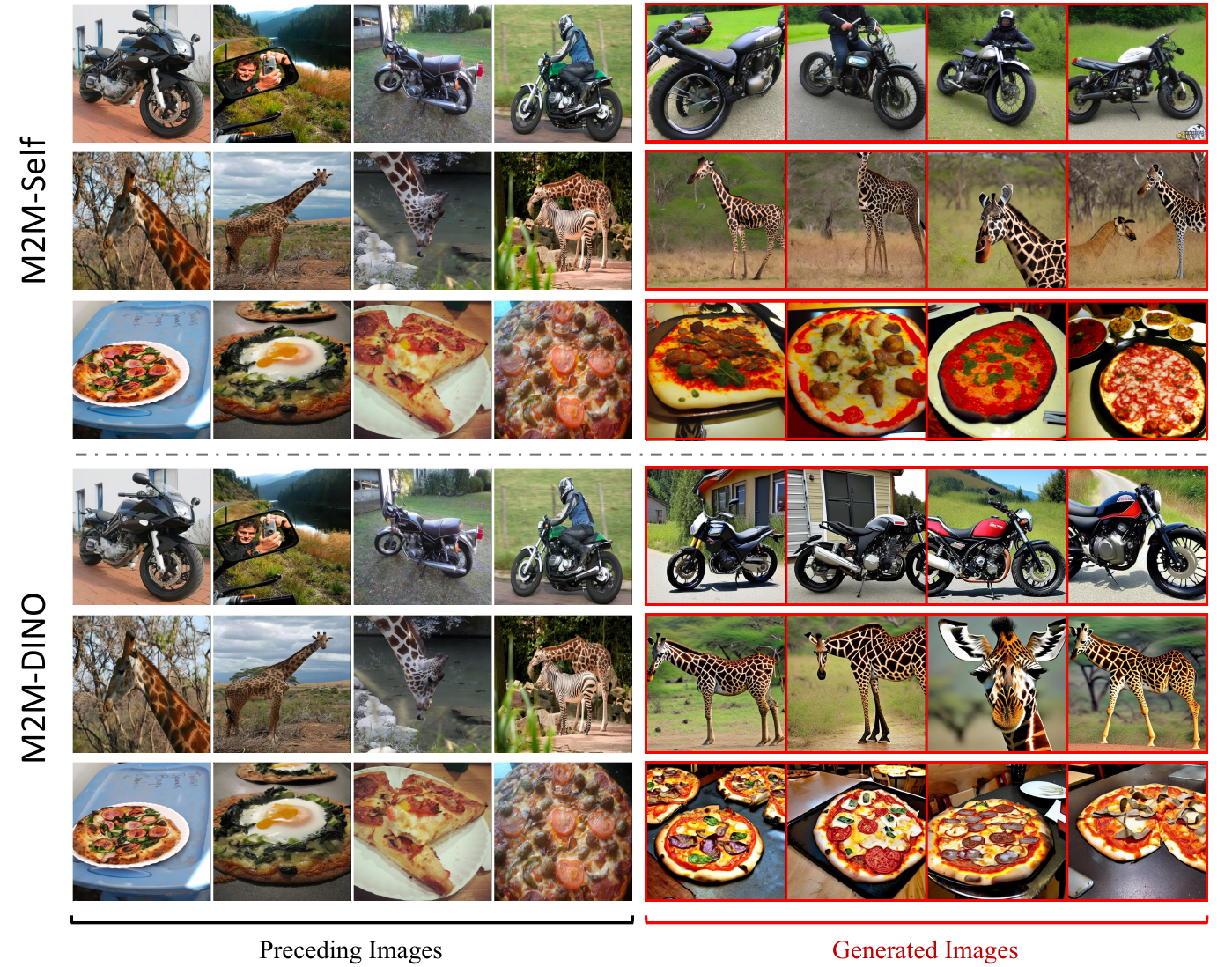}
    \caption{\textbf{Generalization to Real Images.} Columns 1-4 display the real images from the MSCOCO dataset, serving as the preceding images. Columns 5-8 showcase the corresponding images generated by \shortmodelname{}, conditioned on the preceding images. }
    \label{fig:generalization}
    \vspace{-4mm}
\end{figure}
\subsection{Ability to Capture the Relationship/Patterns}

In this section, we investigate the model's capacity to capture the relationship or patterns within preceding images and subsequently generate new images in alignment with the observed patterns. Specifically, we conduct experiments evaluating the following two key aspects: (1) Content consistency and (2) Style consistency. Content consistency evaluates the model's ability to generate images featuring the same type of subject as in the preceding images. Style consistency, on the other hand, examines the model's capability to maintain the aesthetic or stylistic aspects of preceding images, including color schemes, textures, and artistic techniques.

\vspace{-5pt}\paragraph{Content Consistency} In assessing content consistency, we applied \shortmodelname{} to the test subset of \dataname{}. 
As illustrated in Figure \ref{fig:content}, the images generated from \shortsdmodel{} (displayed in the upper section) and \shortdinomodel{} (shown in the bottom section) demonstrate that the proficiency of both model variants in preserving content integrity across a varied range of subjects. From Row 1 to Row 3 for each model variant, we demonstrate that \shortmodelname{} adeptly maintains content consistency in images, starting from simple objects such as a hamburger, progressing to more complex compositions involving multiple objects (like a pinecone, candle, and glass), and extending to detailed indoor environments and scenes involving people.

\vspace{-5pt}\paragraph{Style Consistency}
Style consistency is another critical aspect of our evaluation. As shown in Figure \ref{fig:style}, both \shortsdmodel{} and \shortdinomodel{} effectively replicate a variety of artistic styles derived from preceding images. In each model panel, from Row 1 through Row 3, the model consistently maintains the styles of preceding images, ranging from simplified style to pixel art and watercolor paintings. 

\begin{figure}[!t]
    \centering
\includegraphics[width=0.9\columnwidth]{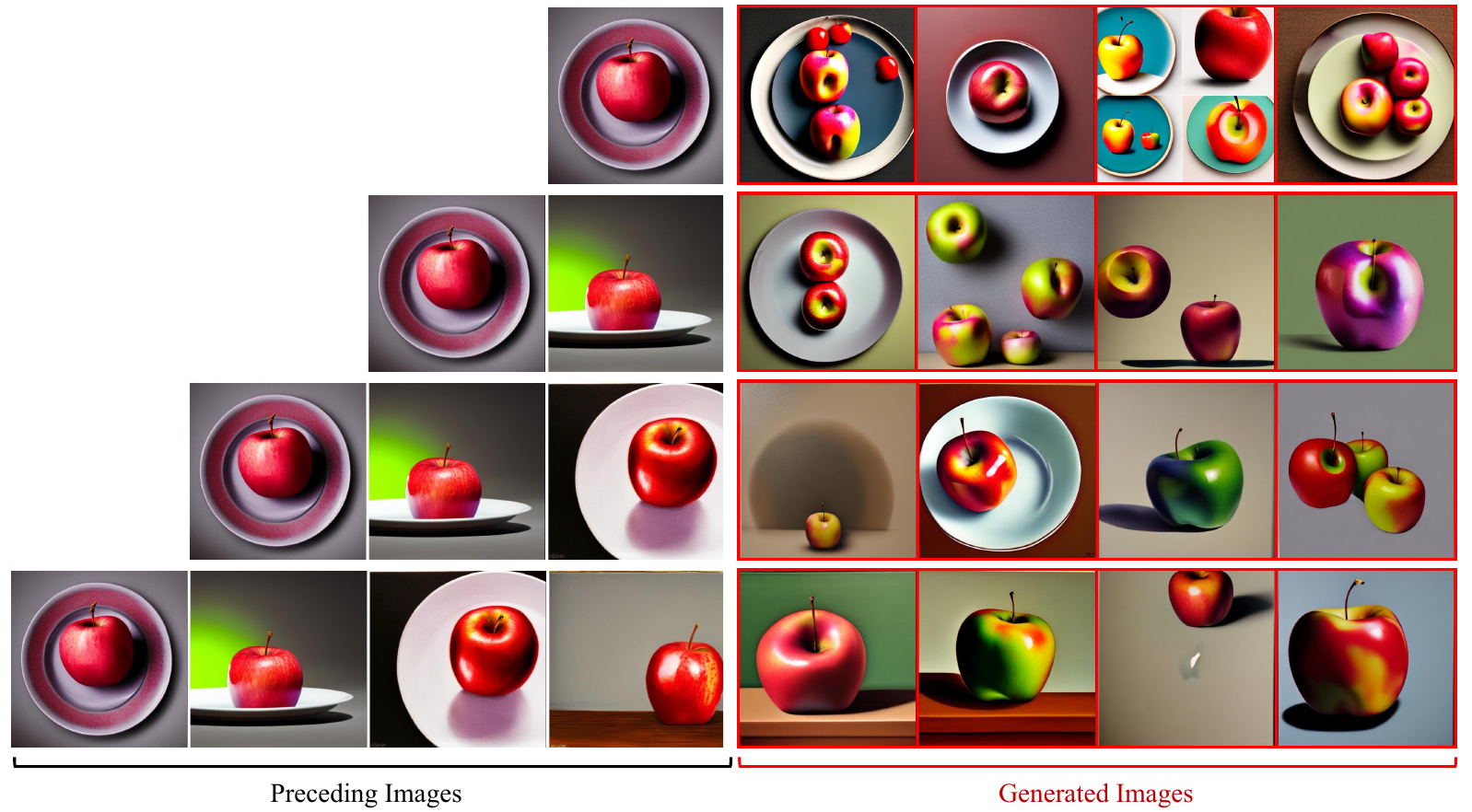}
    \caption{\textbf{Effect of Varying Preceding Images.} The figure presents the images generated from \shortdinomodel{} when conditioned on varying numbers of preceding images. 
    }
    \label{fig:vary_prefix}
    \vspace{-4mm}
\end{figure}

\subsection{Effect of Number of Preceding Images}

We further investigate the effect of altering the numbers of preceding images on multi-image generation. As illustrated in Figure~\ref{fig:vary_prefix}, our experiment involves conditioning \shortdinomodel{} with a different count of preceding images, specifically from one to four, each depicting a single apple. 
Initially, with only one preceding image as condition, \shortdinomodel{} tends to generate images featuring an arbitrary number of apples, diverging from the single-apple pattern. However, as the number of preceding images increases, \shortdinomodel{} begins to more accurately capture and replicate the pattern of a single apple. 
Our findings indicate that as the number of preceding images increases, the images generated by \shortdinomodel{} are more likely to capture and reproduce the patterns observed in preceding images.

\subsection{Generalization to Real Images} We further explore the model's capability for zero-shot generalization to real images, which is pivotal in understanding how well \shortmodelname{} can adapt to real-world scenarios beyond the synthetic data it was trained on. For this purpose, we employ MSCOCO \cite{lin2014microsoft} dataset, which contains real images of complex everyday scenes containing common objects. 
To assess the model's capability in maintaining content consistency across various real-world scenarios, we group images from MSCOCO into sets based on object categories. Each set contained different images, but all shared the same object category. 

Figure \ref{fig:generalization} showcases images generated (Columns 5-8) by \shortsdmodel{} and \shortdinomodel{}, which are based on real images from the MSCOCO (Columns 1-4). 
Impressively, despite being trained solely on synthetic data, our model exhibits zero-shot generalization to \textit{real} images. The generated images not only resemble the real images but also maintain a high degree of content consistency.

\begin{table}[!t]
  \centering
  \resizebox{\linewidth}{!}{%
  \begin{tabular}{l | c c c c}
  \toprule
   \textbf{Method} & FID $\downarrow$  & IS $\uparrow$  & Text-Image CLIP $\uparrow$ & Image-Image CLIP  $\uparrow$ \\
  \midrule

  \shortsdmodel{} (9M) &  9.56 $\pm$ 1.21 & 26.19 $\pm$ 0.67 & 22.71 $\pm$ 0.52 & 76.29 $\pm$ 0.02 \\
  \shortdinomodel{} (6M) & 8.88 $\pm$ 0.87 & 28.07 $\pm$ 0.58 & 23.05 $\pm$ 0.49 & 77.41 $\pm$ 0.03  \\
    
  \bottomrule
  \end{tabular}}
  \caption{\textbf{Quantitative Evaluation on 10K \dataname{} Test Subset.} Each metric is reported as an average score $\pm$ standard deviation across the 10 generated images.}
  \label{table:score}
  \vspace{-6mm}
  \end{table}

\subsection{Quantitative Evaluation}

In this section, we present a comprehensive quantitative assessment of \shortmodelname{}. Our focus is twofold: to analyze the quality of the images generated by the model, and to determine its effectiveness in producing images that are visually consistent with a given sequence of preceding images.
Our evaluation leverages three established metrics: Fréchet Inception Distance (FID), Inception Score (IS), and various CLIP scores, utilizing a randomly chosen subset of 10,000 samples from the \dataname{} test split.

\shortmodelname{} is designed to accept a sequence of preceding images and autoregressively generate subsequent images. For this analysis, we generated 10 images using both \shortsdmodel{} and \shortdinomodel{}, conditioned on four preceding images. The FID, IS, and CLIP scores were computed for every $n$-th image generated.
FID is calculated by measuring the Fréchet distance between the multivariate Gaussian distributions of the `real' (the first preceding images) and the generated images. 
To assess the consistency of the generated images with the preceding ones, we employ two variants of CLIP scores: text-image and image-image. The Text-Image CLIP score is calculated by comparing each $n$-th generated image to the common textual description associated with the preceding images in \dataname{}. The Image-Image CLIP score measured the visual similarity of each $n$-th generated image with all preceding images.
Table~\ref{table:score} details the average scores and standard deviations for these metrics, reported as an average score $\pm$ standard deviation across 10 generated images. Notably, \shortdinomodel{} outperforms \shortsdmodel{} across all metrics, indicating a more robust capability in generating high-quality and contextually consistent images within a sequence.

\begin{figure}[!t]
    \centering
\includegraphics[width=0.9\linewidth]{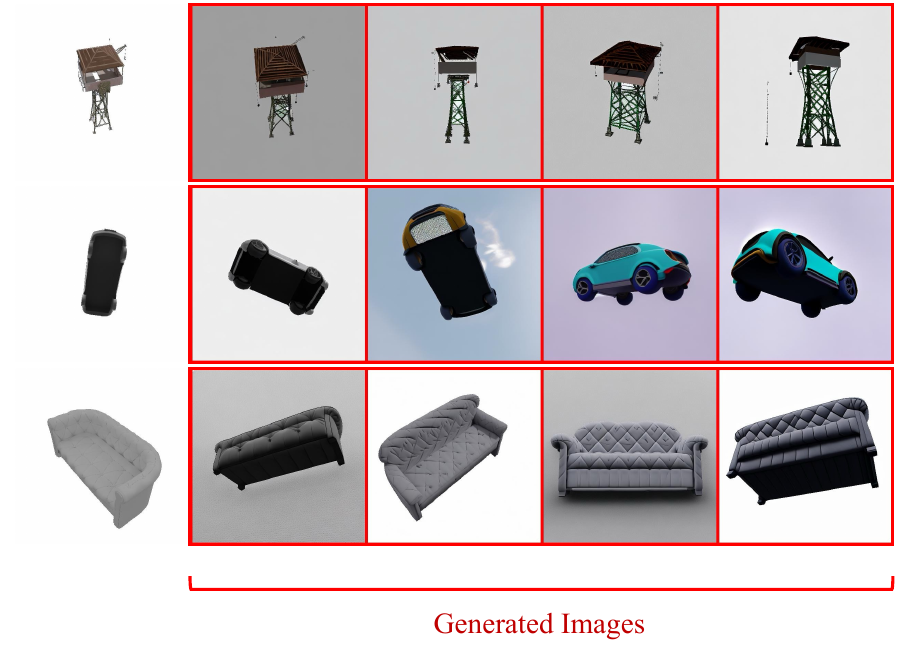}
    \caption{\textbf{Novel View Synthesis on Objaverse} Column 1 presents a singular preceding image of an object. Columns 2 - 5 display the images generated by \shortdinomodel{} from various novel viewpoints.}
    \label{fig:obj}
    \vspace{-6mm}
\end{figure}

\subsection{Adaptation for Various Multi-Image Tasks}

In this section, we present \shortmodelname{} fine-tuning results on two different multi-image generation tasks, demonstrating its versatility and effectiveness across different applications.

\begin{figure}[!t]
    \centering
\includegraphics[width=\linewidth]{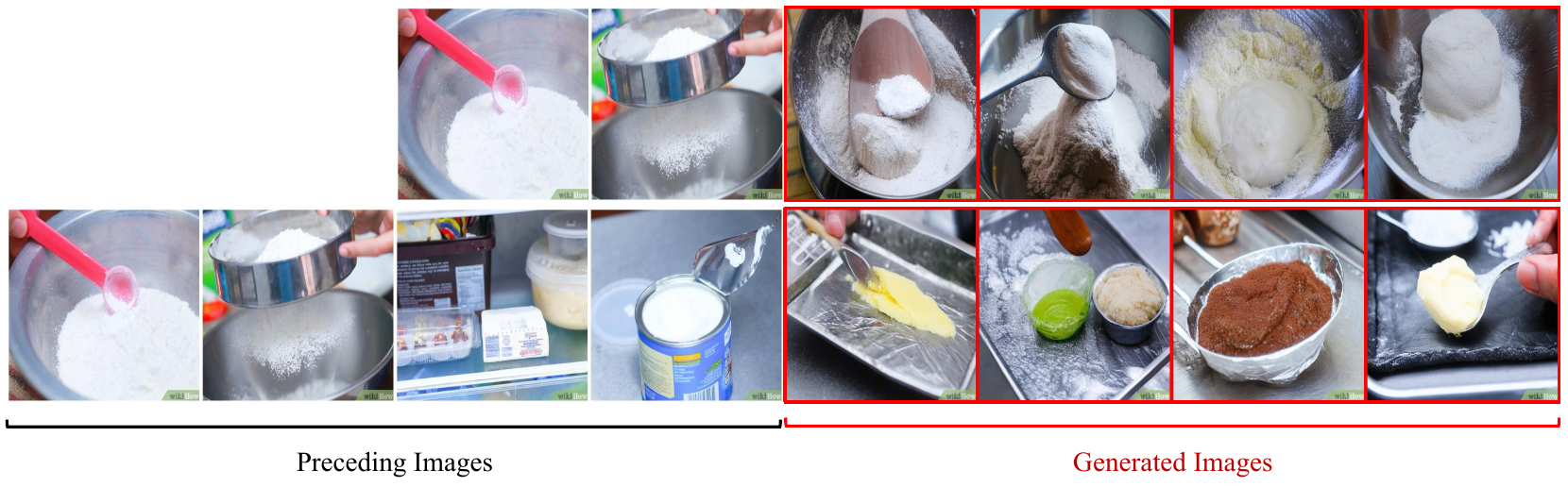}
    \caption{\textbf{Visual Procedure Generation on VGSI} 
    Columns 1-4: the sequence of historical visual steps. Columns 5 - 8: images generated by \shortdinomodel{} that depict future visual steps.
}
    \label{fig:vgsi}
    \vspace{-4mm}
\end{figure}

\subsubsection{Novel-View Synthesis}

The Novel-View Synthesis task tests the model's capability to generate images from new viewpoints. This is achieved by integrating a camera embedding that encodes information about the camera's viewpoint. Specifically, this embedding captures the camera extrinsic for each image using a straightforward Multilayer Perceptron (MLP) layer.
Figure~\ref{fig:obj} demonstrates the images generated by \shortdinomodel{} when conditioned on a singular image. We show that \shortdinomodel{} is capable of auto-gressively generating multi-view images that are consistent with each other when conditioned on one initial preceding image.



\subsubsection{Visual Procedure Generation}

The Visual Procedure Generation task challenges \shortmodelname{} to understand and predict the sequence of visual steps in a procedure. To equip the model with the necessary understanding of sequence and progression, we introduce a positional embedder. This embedder utilizes sinusoidal encoding to capture the position of each image in the sequence, which is then processed through an MLP layer.
Figure~\ref{fig:vgsi} showcases the model's ability to predict future steps in a visual procedure based solely on the sequence of input images.


\subsection{Sampling Efficiency}

\begin{figure}[!t]
    \centering
    \includegraphics[width=0.92\linewidth]{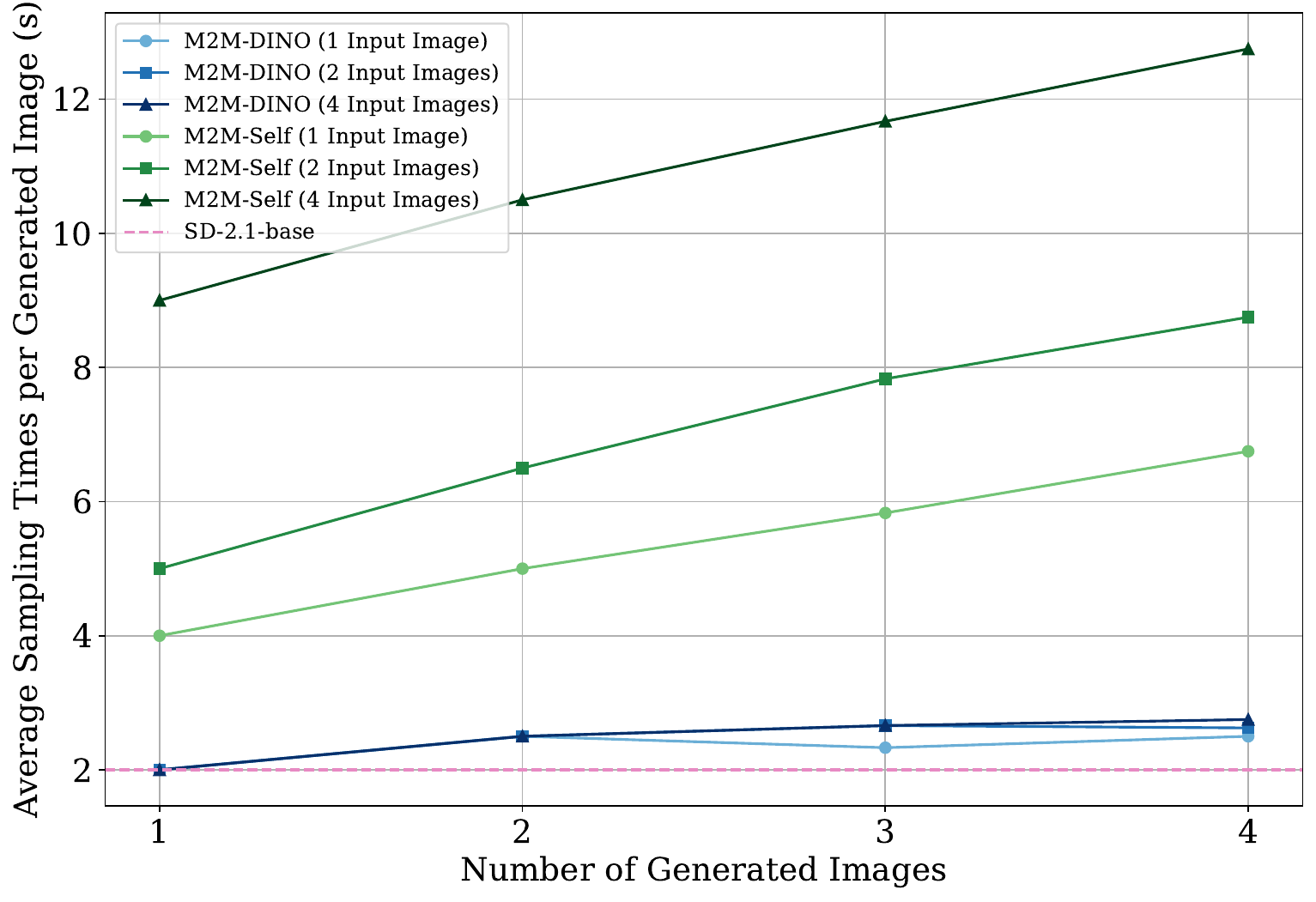}
    \caption{\textbf{Sampling Efficiency.} 
    The sampling speed is measured as the average time to generate one image when using the DDIM sampler with 50 denoising steps on a single NVIDIA A40 GPU.
    The efficiency is measured across \shortsdmodel{} and \shortdinomodel{}, when using 1, 2, and 4 input images, and compared against the StableDiffusion-2.1-base.
    }
    \label{fig:sampling_speed}
    \vspace{-4mm}
\end{figure}

We evaluate the sampling efficiency of our proposed \shortdinomodel{} and \shortsdmodel{} methodologies against the Stable Diffusion-2.1-base (SD-2.1-base).
Figure \ref{fig:sampling_speed} illustrates the comparative analysis of sampling speed, measured as the average time required to generate a single image, across different model configurations, considering various numbers of input and generated images. All the models utilize the DDIM sampler with 50 denoising steps, and the evaluation is performed on a single NVIDIA A40 GPU to ensure a fair and consistent basis for comparison. 

Our results indicate that \shortdinomodel{} significantly outperforms \shortsdmodel{} in terms of sampling efficiency, especially as the number of input and generated images rises. 
Notably, \shortdinomodel{} demonstrates a sampling speed on par with SD-2.1-base, even when \shortdinomodel{} is set to process multiple input images and generate multiple images. 
These results highlight the robust and consistent efficiency of \shortdinomodel{} in many-to-many image generation.

\section{Limitations}
\label{sec:limitations}

While our model achieves considerable success in multi-image generation, it is not without its limitations. Notably, it struggles to generate human faces with high fidelity, a shortfall possibly stemming from the suboptimal quality of human faces present in our synthetic training set. Future efforts could benefit from incorporating more advanced diffusion models to enhance the quality of training data, particularly for human faces.

Another observed challenge is the gradual decline in image quality during the auto-regressive generation of prolonged image sequences. This performance degradation highlights a potential area for further optimization, suggesting a need for improved strategies to maintain image quality throughout extended generative processes, which is critical for applications requiring the continuous production of images.

\section{Conclusion}
\label{conclusion}

We introduce \dataname{}, a novel large-scale multi-image dataset, containing 12M synthetic multi-image samples, each with 25 interconnected images. We propose a domain-general \longmodelname{} (\shortmodelname{}) model that can perceive and generate an arbitrary number of interrelated images auto-regressively. 
We explore two main model variants, \shortsdmodel{} and \shortdinomodel{}, both demonstrating exceptional ability in capturing and replicating style and content from preceding images when trained on \dataname{}.
Remarkably, our model exhibits zero-shot generalization to \textit{real} images despite being trained solely on synthetic data. We further demonstrate the model’s adaptability to various multi-image generation tasks, including Novel View Synthesis and Visual Procedure Generation, through targeted fine-tuning, underscoring the potential of our approach to adapt to a broad spectrum of multi-image generation tasks. 
\section*{Impact Statement}

This paper presents work whose goal is to advance the field of Machine Learning. There are many potential societal consequences of our work, none which we feel must be specifically highlighted here.


\bibliography{example_paper}
\bibliographystyle{icml2024}

\newpage
\appendix
\onecolumn

\section{Implementation Details}

For our training procedure, we adopt two configurations for \sdmodel{} (\shortsdmodel{}) and \dinomodel{} (\shortdinomodel{}).
Both \shortsdmodel{} and \shortdinomodel{} are trained with a total batch size of 256 on 8 $\times$ 80GB NVIDIA A100 GPUs for one epoch. We use a learning rate of $10^{-5}$ without any learning rate warm-up. \modelname{} is initialized from the EMA weights of the Stable Diffusion v2-1 base \footnote{https://huggingface.co/stabilityai/stable-diffusion-2-1-base}. For other configurations, we adopt the default training settings provided within the Stable Diffusion codebase. 
Regarding the encoding of preceding images in \dinomodel{}, we employ DINOv2-giant~\footnote{https://huggingface.co/facebook/dinov2-giant}, particularly leveraging its last hidden states.
At inference time, \modelname{} generates novel images with 50 denoising steps using the DDIM sampler~\cite{song2020denoising}. A guidance scale of $7.5$ is employed for the preceding images unless specified otherwise.


\section{Additional Experiments}

\subsection{Generation of Images Conditioned on Synthetic Images}

Figures \ref{fig:syn_self_1}, \ref{fig:syn_self_2}, and \ref{fig:syn_self_3} showcase images generated by the \shortsdmodel{}. On the other hand, Figures \ref{fig:syn_dino_1}, \ref{fig:syn_dino_2}, and \ref{fig:syn_dino_3} display the images generated from the \shortdinomodel{}. These images were generated using the conditioning images from the test subset of the \dataname{} dataset.

\begin{figure}[!t]
    \centering
    \includegraphics[width=\linewidth]{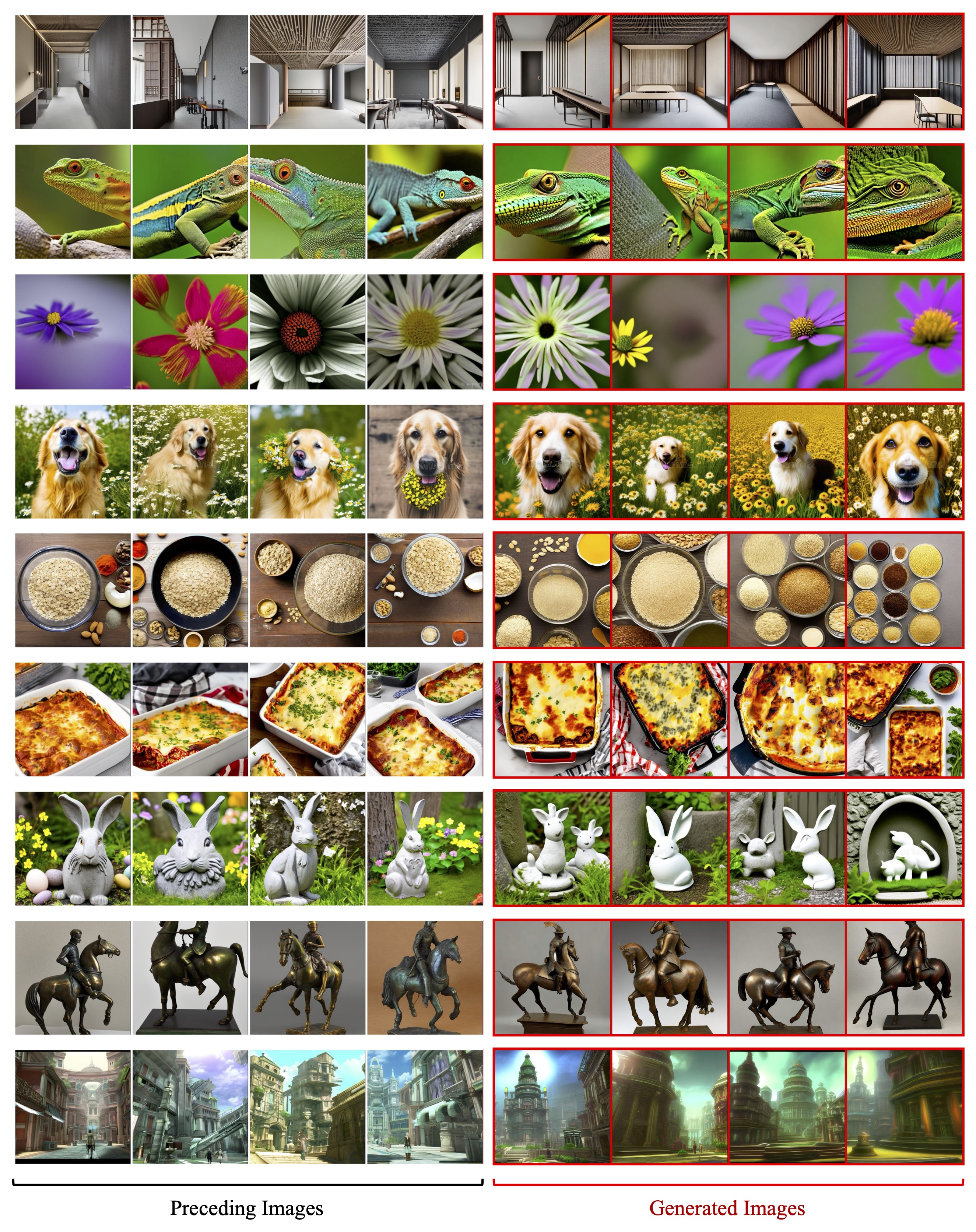}
    \caption{\textbf{Images generated by \shortsdmodel{}.} Columns 1-4 showcase the preceding images for conditioning, and Columns 5-8 display the generated images. }
    \label{fig:syn_self_1}
\end{figure}

\begin{figure}[!t]
    \centering
    \includegraphics[width=\linewidth]{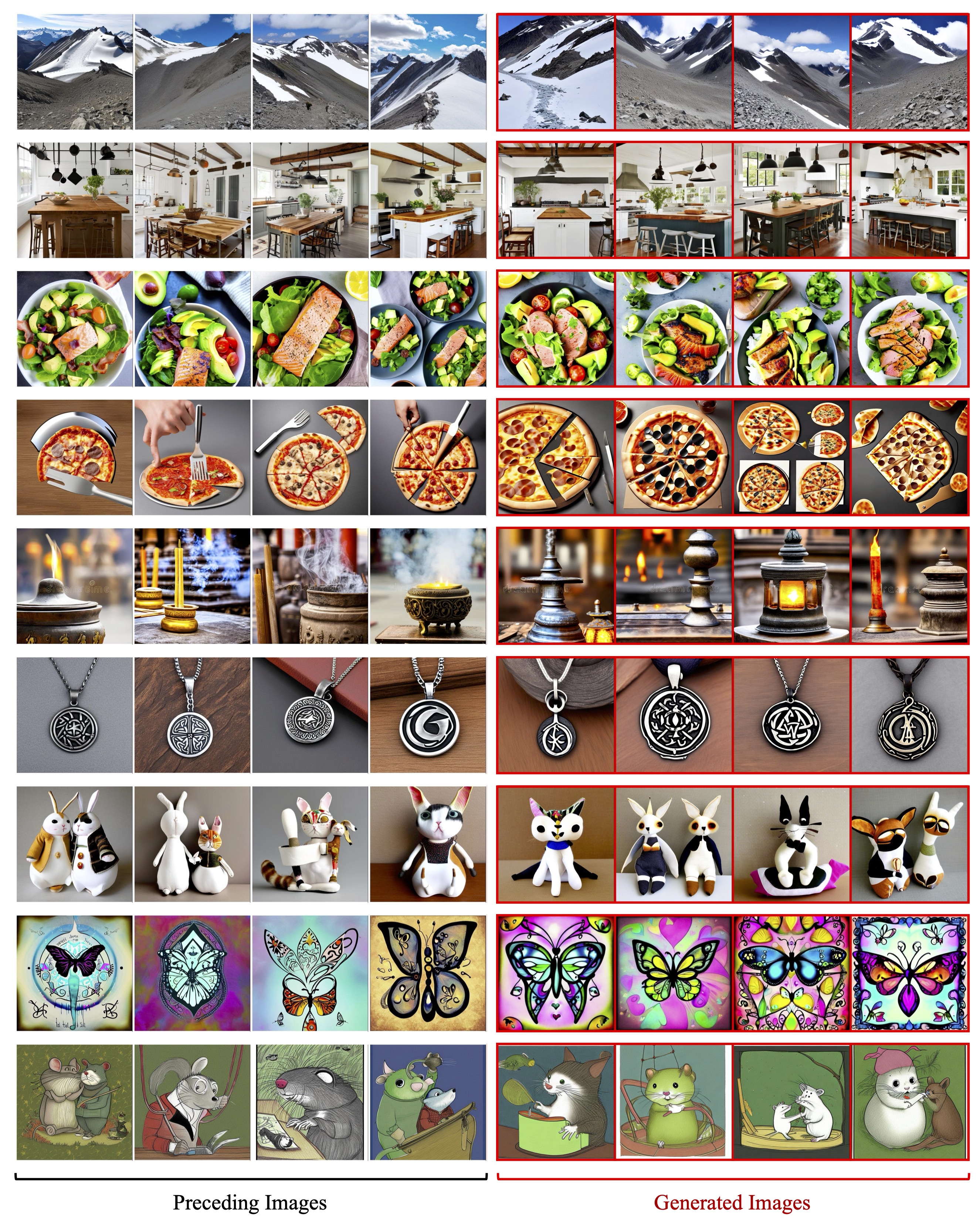}
    \caption{\textbf{(Continued) Images generated by \shortsdmodel{}.} Columns 1-4 showcase the preceding images for conditioning, and Columns 5-8 display the generated images. }
    \label{fig:syn_self_2}
\end{figure}

\begin{figure}[!t]
    \centering
    \includegraphics[width=\linewidth]{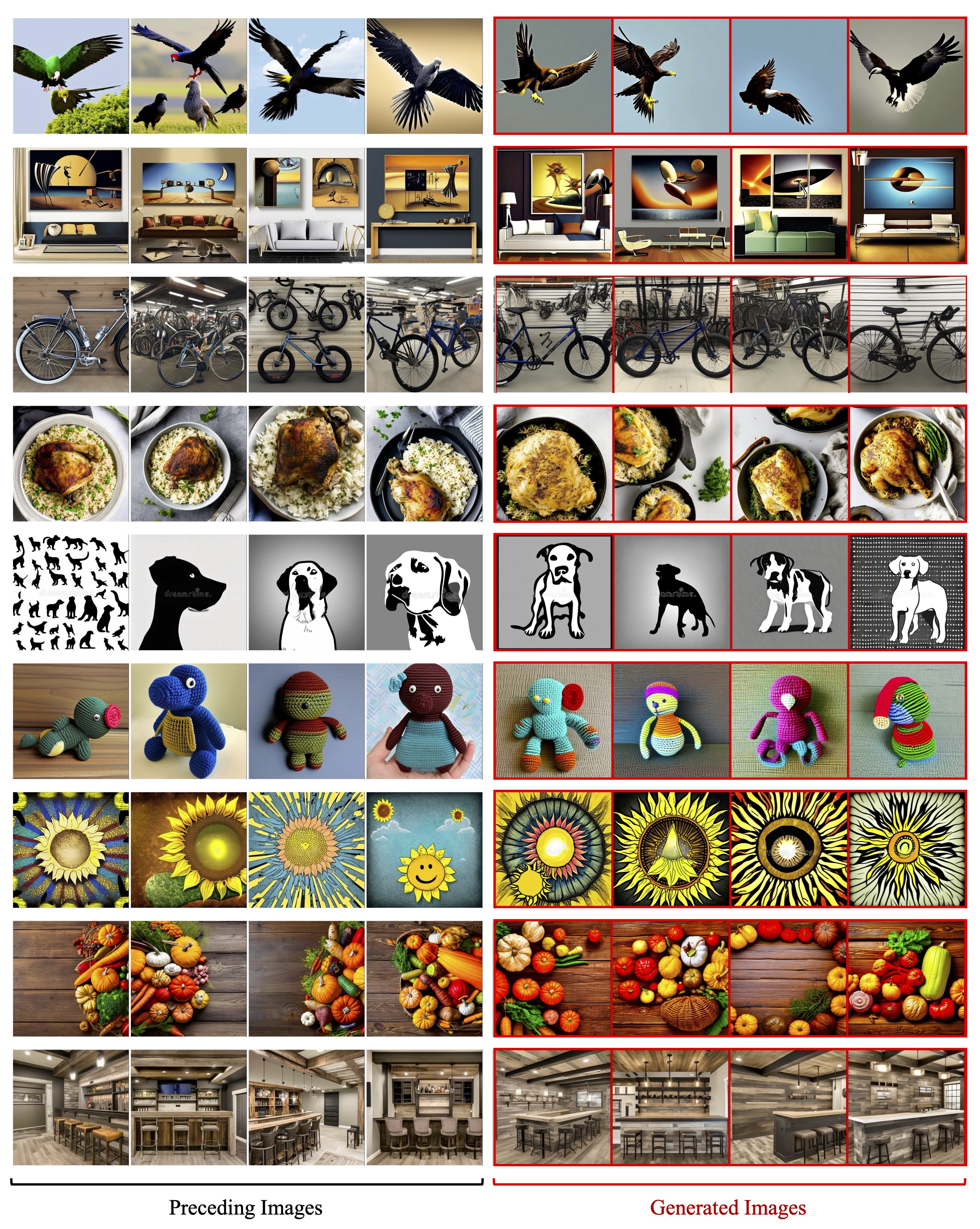}
    \caption{\textbf{(Continued) Images generated by \shortsdmodel{}.} Columns 1-4 showcase the preceding images for conditioning, and Columns 5-8 display the generated images. }
    \label{fig:syn_self_3}
\end{figure}
\begin{figure}[!t]
    \centering
    \includegraphics[width=\linewidth]{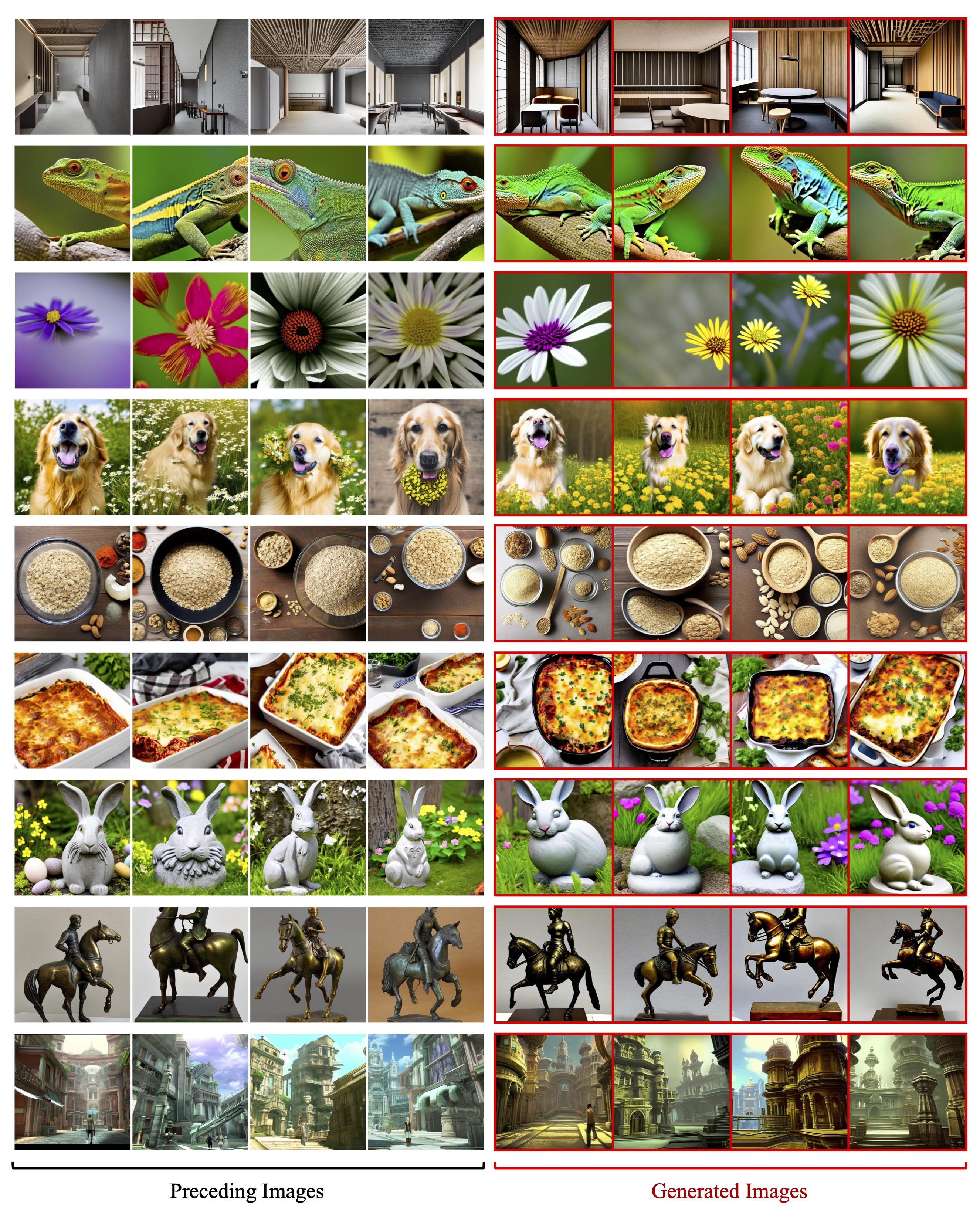}
    \caption{\textbf{Images generated by \shortdinomodel{}.} Columns 1-4 showcase the preceding images for conditioning, and Columns 5-8 display the generated images. }
    \label{fig:syn_dino_1}
\end{figure}

\begin{figure}[!t]
    \centering
    \includegraphics[width=\linewidth]{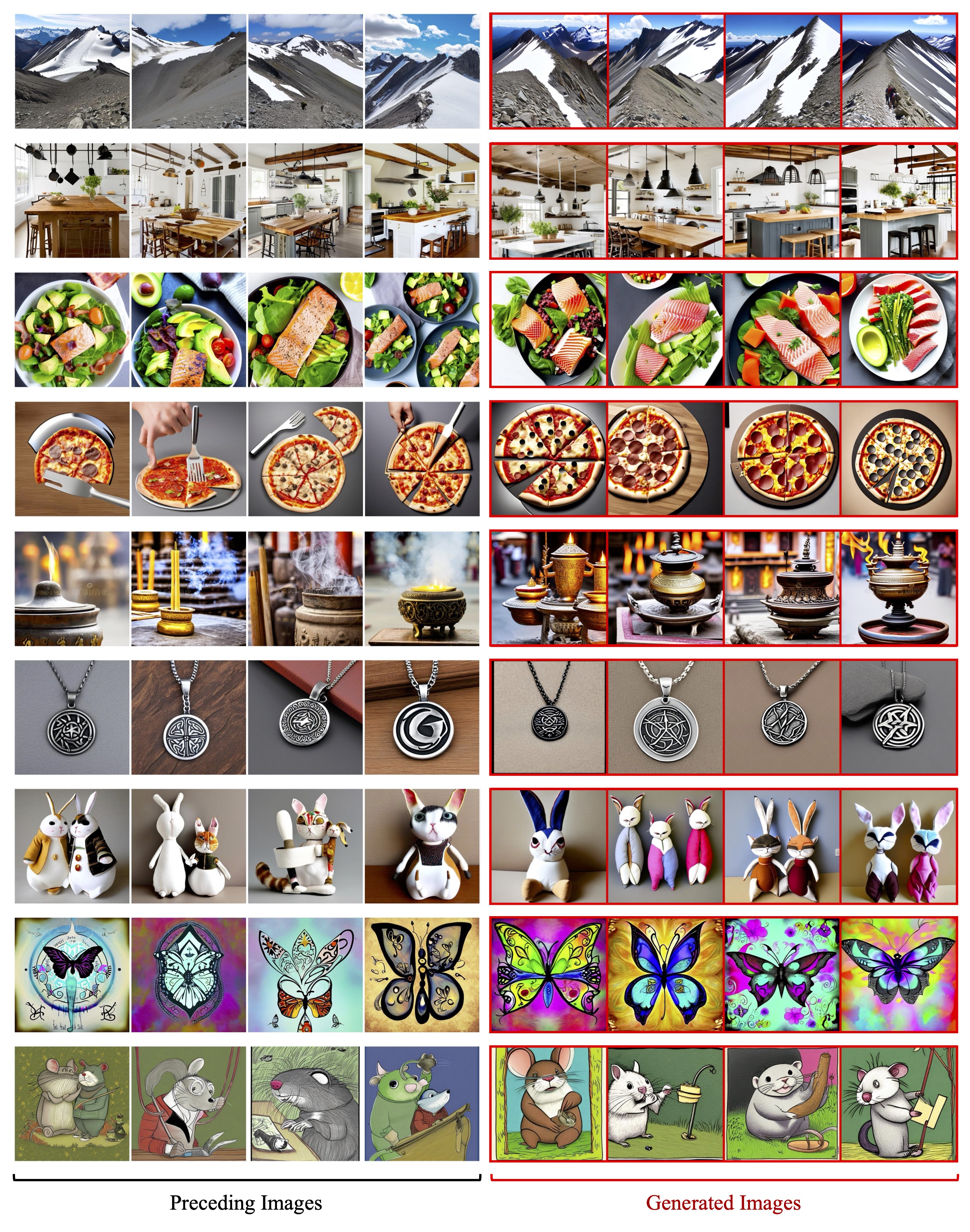}
    \caption{\textbf{(Continued) Images generated by \shortdinomodel{}.} Columns 1-4 showcase the preceding images for conditioning, and Columns 5-8 display the generated images. }
    \label{fig:syn_dino_2}
\end{figure}

\begin{figure}[!t]
    \centering
    \includegraphics[width=\linewidth]{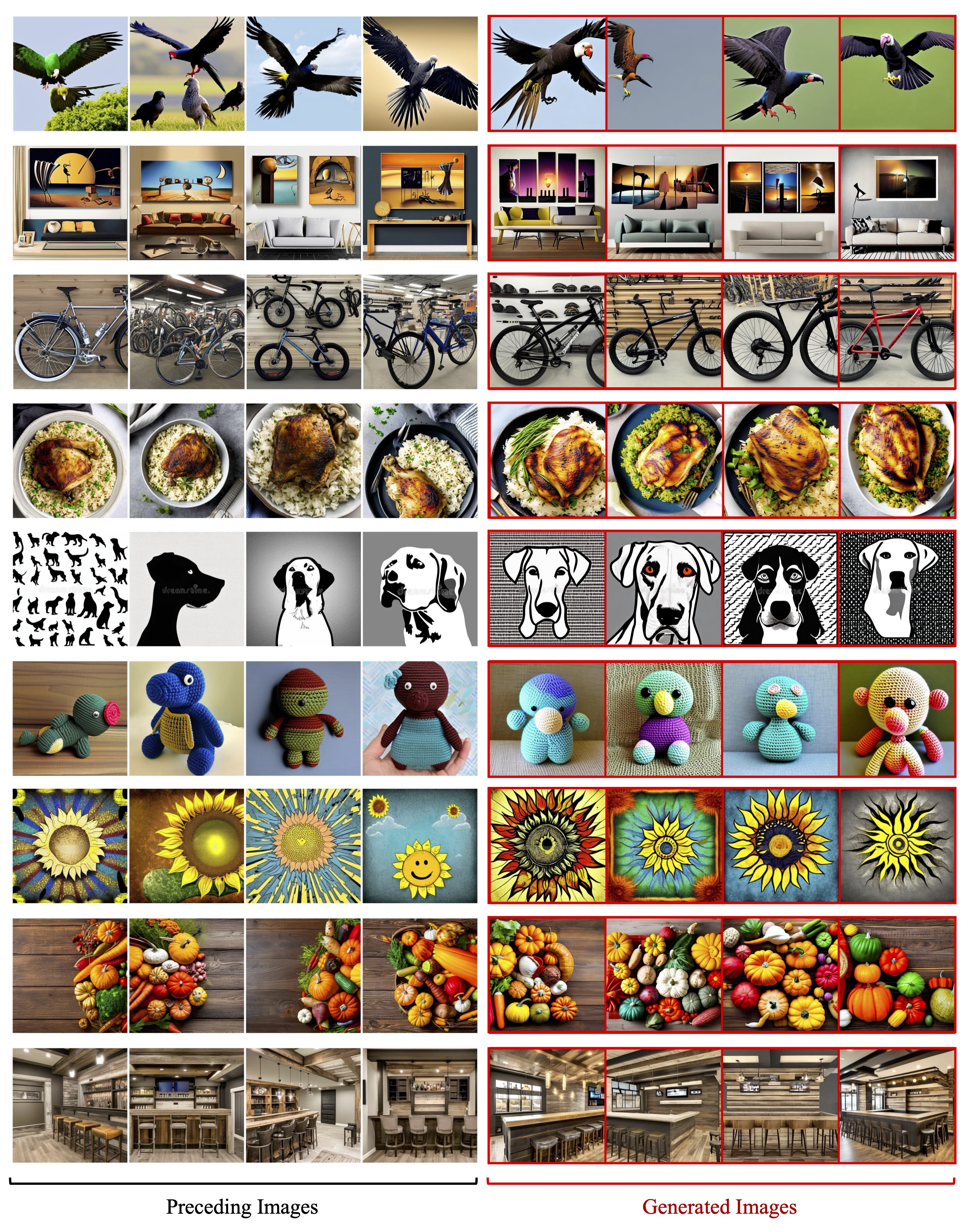}
    \caption{\textbf{(Continued) Images generated by \shortdinomodel{}.} Columns 1-4 showcase the preceding images for conditioning, and Columns 5-8 display the generated images. }
    \label{fig:syn_dino_3}
\end{figure}

\subsection{Generation of Images Conditioned on Real Images}

\begin{figure}[!t]
    \centering
    \includegraphics[width=\linewidth]{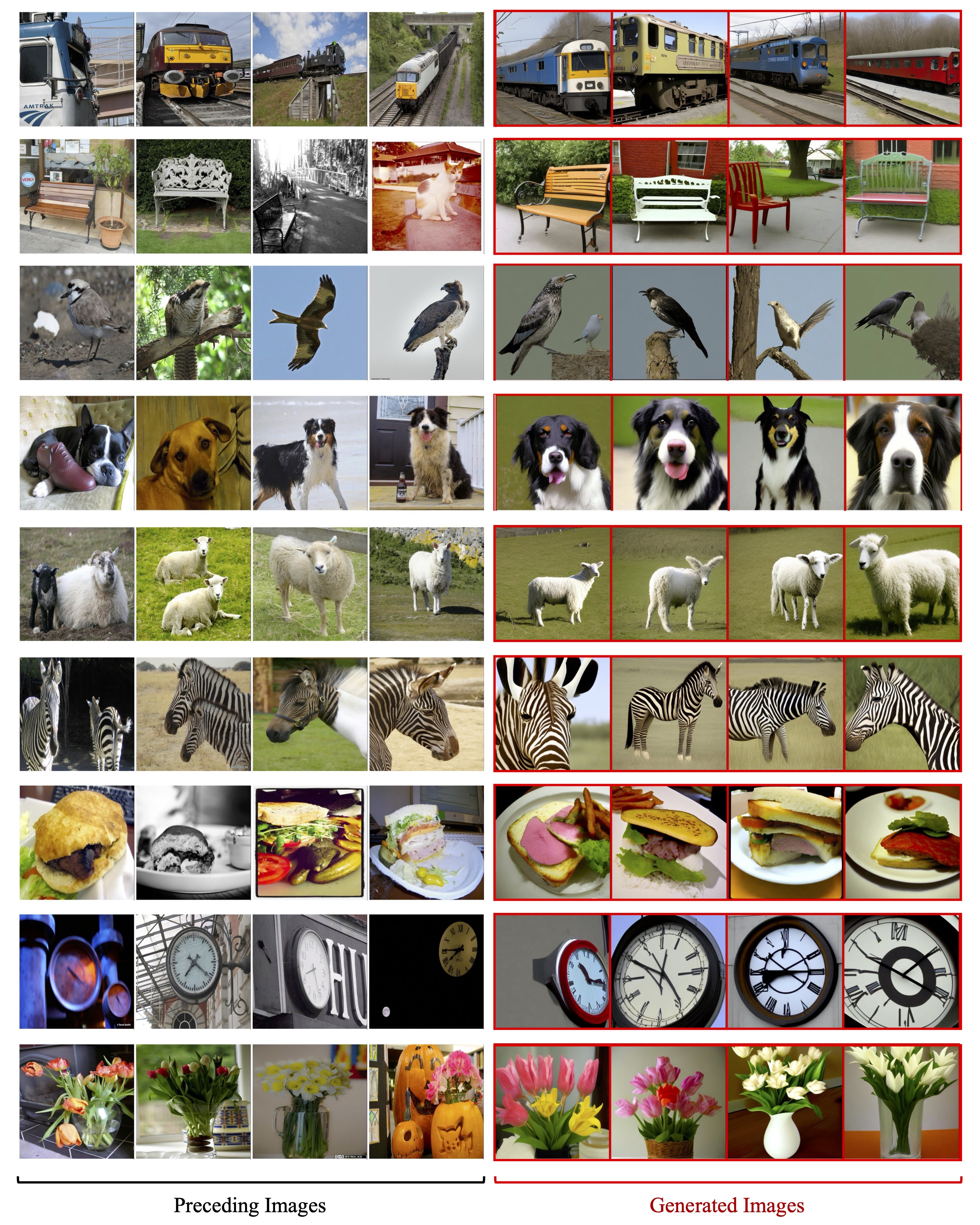}
    \caption{\textbf{Generalization to Real Images.} Columns 1-4 display the real images from the MSCOCO dataset, serving as the preceding images. Columns 5-8 showcase the corresponding images generated by \textbf{\shortsdmodel{}}, conditioned on the preceding images. }
    \label{fig:mscoco_self}
    \vspace{-4mm}
\end{figure}

\begin{figure}[!t]
    \centering
    \includegraphics[width=\linewidth]{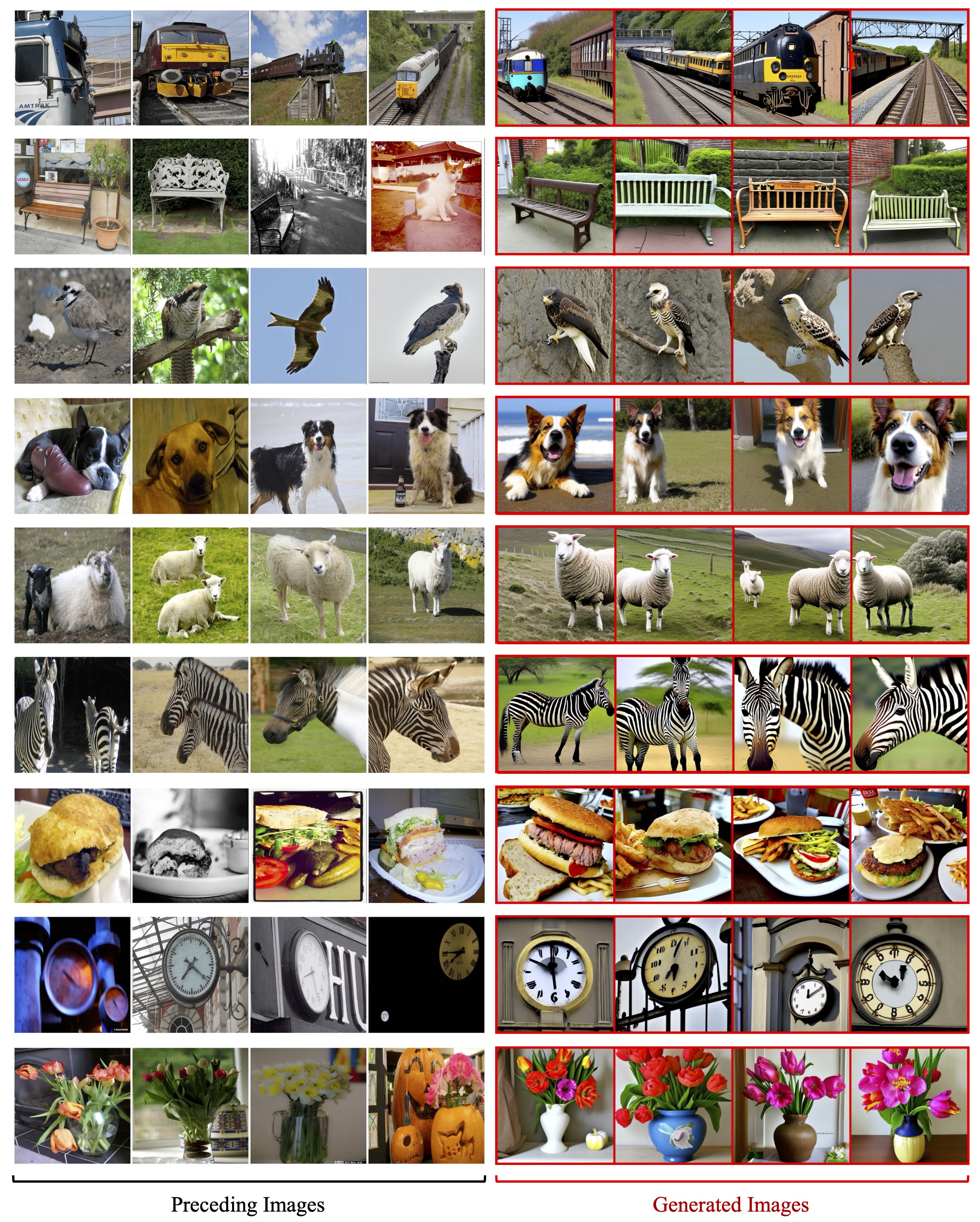}
    \caption{\textbf{Generalization to Real Images.} Columns 1-4 display the real images from the MSCOCO dataset, serving as the preceding images. Columns 5-8 showcase the corresponding images generated by \textbf{\shortdinomodel{}}, conditioned on the preceding images. }
    \label{fig:mscoco_dino}
    \vspace{-4mm}
\end{figure}

Figure \ref{fig:mscoco_self} and \ref{fig:mscoco_dino} present a comparative showcase of images generated by \shortsdmodel{} and \shortdinomodel{}, respectively. These images are generated from real-world scenes contained in the MSCOCO dataset. The first four columns (Columns 1-4) display the original images from the MSCOCO dataset, whereas the generated images are presented in the subsequent columns (Columns 5-8).


\end{document}